\documentclass[letterpaper, 10 pt, conference]{ieeeconf}  

\IEEEoverridecommandlockouts                              

\usepackage{graphicx}
\usepackage{wrapfig}
\usepackage{adjustbox}
\usepackage{booktabs}
\usepackage{multirow}
\usepackage{amsmath}
\usepackage{dsfont}
\usepackage{amssymb}
\usepackage{pifont} 
\usepackage[table,xcdraw]{xcolor}
\newcommand{\cmark}{\ding{51}} %
\newcommand{\xmark}{\ding{55}} %

\usepackage[pagebackref,breaklinks,colorlinks]{hyperref}

\usepackage[capitalize]{cleveref}
\crefname{section}{Sec.}{Secs.}
\Crefname{section}{Section}{Sections}
\Crefname{table}{Table}{Tables}
\crefname{table}{Tab.}{Tabs.}

\begin{document}
\title{Diving into the Depths of Spotting Text in Multi-Domain Noisy Scenes}

\author{Alloy Das\textsuperscript{\textsection}$^{1}$, Sanket Biswas\textsuperscript{\textsection}$^{2}$, Umapada Pal$^{1}$ and Josep Llad\'{o}s$^{2}$
\thanks{$^{1}$Alloy Das and Umapada Pal are with Computer Vision and Pattern Recognition Unit, Indian Statistical Institute, Kolkata, India
       {\tt\small \{alloydas\_t, umapada\}@isical.ac.in}}%
\thanks{$^{2}$Sanket Biswas and Josep Llad\'{o}s are with the Computer Vision Center, Computer Science Department, Universitat Aut\'{o}noma de Barcelona, Barcelona, Spain
        {\tt\small \{sbiswas, josep\}@cvc.uab.es}}%
}



\maketitle
\begingroup\renewcommand\thefootnote{\textsection}
\footnotetext{Equal contribution}
\endgroup
\begin{abstract}
When used in a real-world noisy environment, the capacity to generalize to multiple domains is essential for any autonomous scene text spotting system. However, existing state-of-the-art methods employ pretraining and fine-tuning strategies on natural scene datasets, which do not exploit the feature interaction across other complex domains. In this work, we explore and investigate the problem of \textit{domain-agnostic scene text spotting}, i.e., training a model on multi-domain source data such that it can directly generalize to target domains rather than being specialized for a specific domain or scenario. In this regard, we present the community a text spotting validation benchmark called \textit{Under-Water Text (UWT)} for noisy underwater scenes to establish an important case study. Moreover, we also design an efficient super-resolution based end-to-end transformer baseline called \textit{DA-TextSpotter} which achieves comparable or superior performance over existing text spotting architectures for both regular and arbitrary-shaped scene text spotting benchmarks in terms of both accuracy and model efficiency. The dataset, code and pre-trained models will be released upon acceptance. 

\end{abstract}

\section{Introduction}
\begin{figure}[ht]
\centering
\begin{tabular}{c c c } 
\includegraphics[height=2cm, width= 2cm]{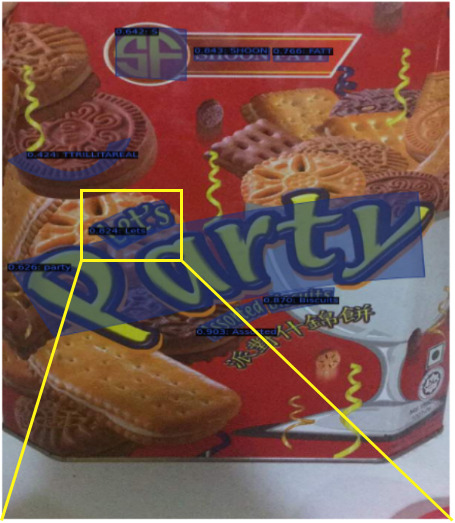}&
\includegraphics[height=2cm, width= 2cm]{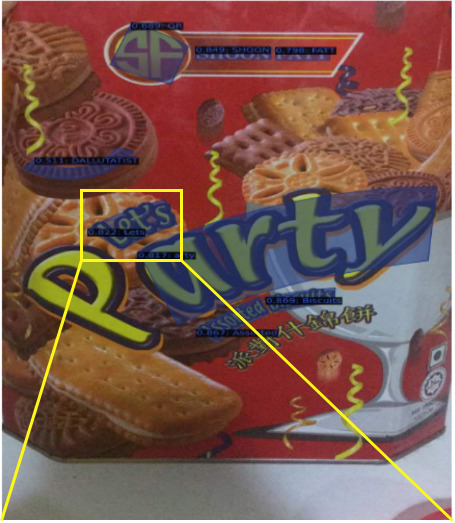}&
\includegraphics[height=2cm, width= 2cm]{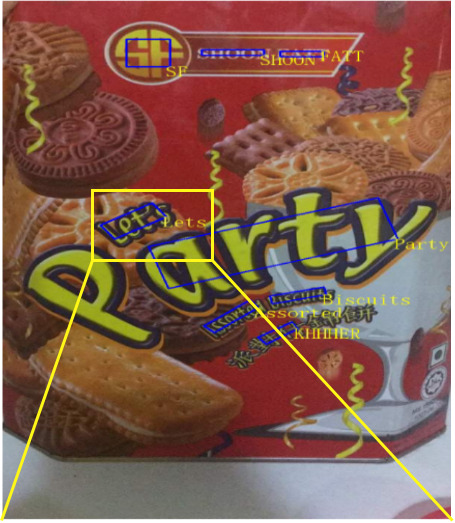}\\
\includegraphics[height=2cm, width= 2cm]{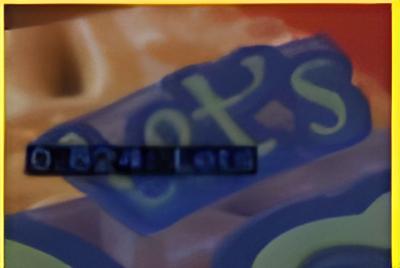}&
\includegraphics[height=2cm, width= 2cm]{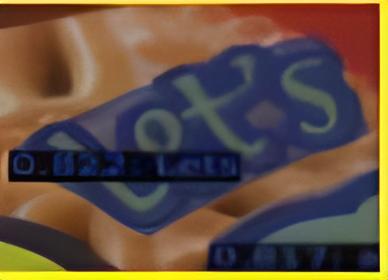}&
\includegraphics[height=2cm, width= 2cm]{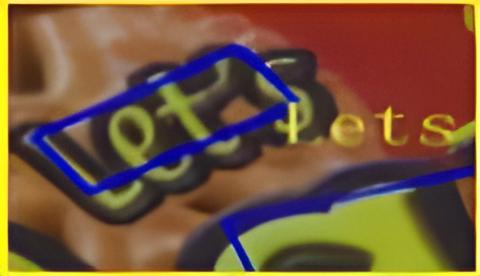}\\
(a) ABCNet &(b) ABCNetv2 &(c) MANGO \\
\includegraphics[height=2cm, width= 2cm]{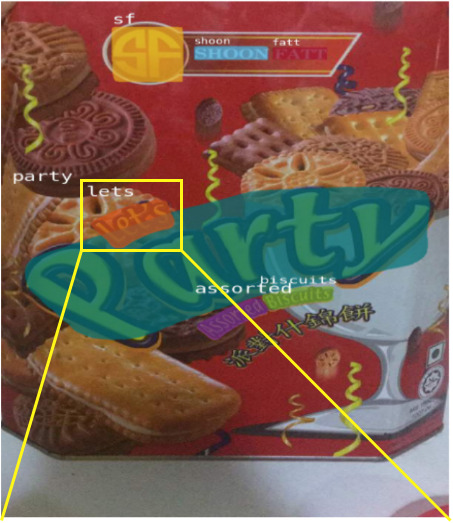}&
\includegraphics[height=2cm, width= 2cm]{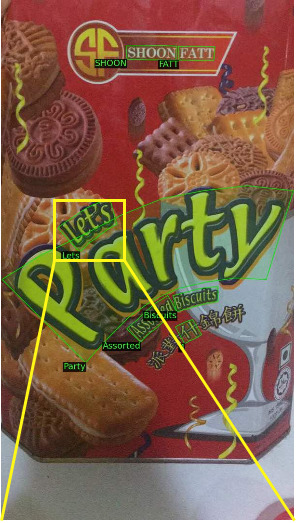}&
\includegraphics[height=2cm, width= 2cm]{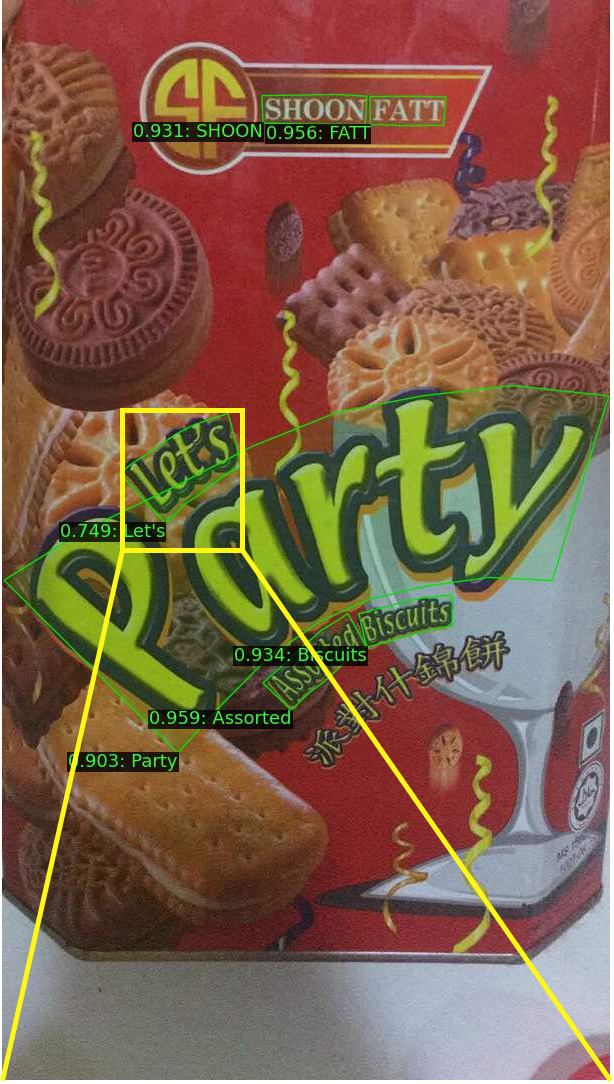}\\
\includegraphics[height=2cm, width= 2cm]{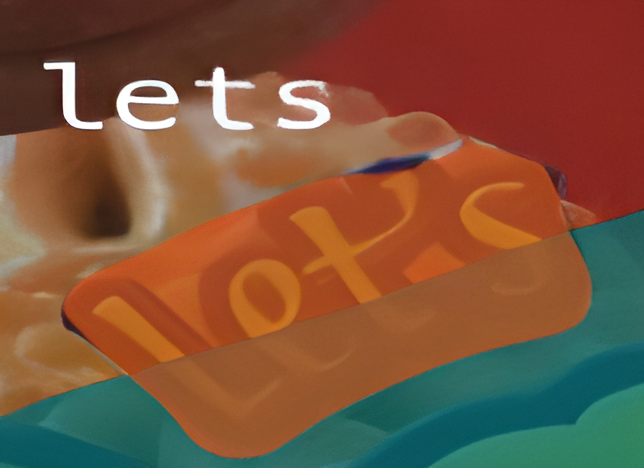}&
\includegraphics[height=2cm, width= 2cm]{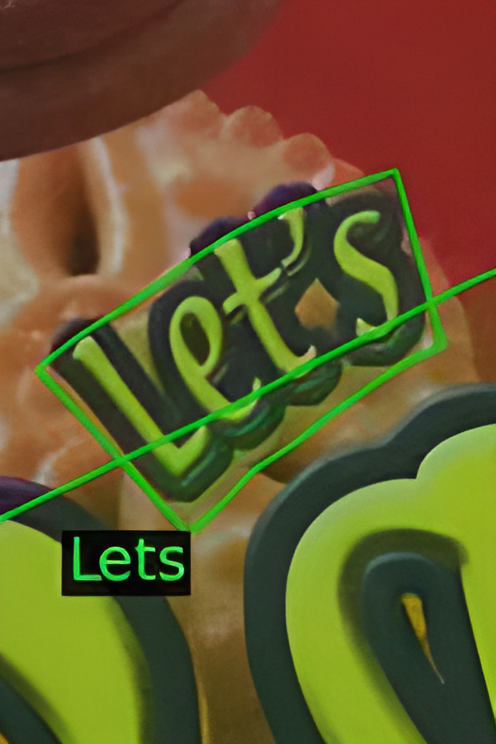}&
\includegraphics[height=2cm, width= 2cm]{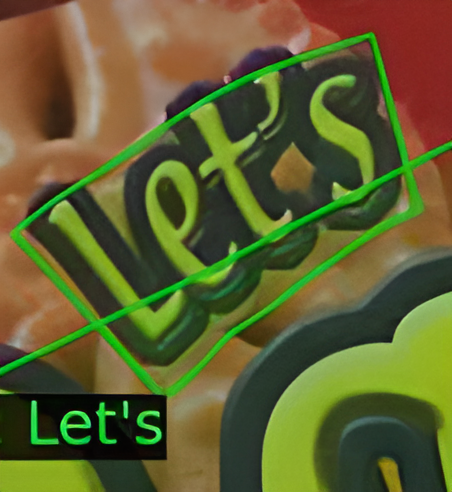}\\
(d) SwinTextSpotter &(e) TESTR & (f) \textbf{DA-TextSpotter}\\
\end{tabular}
\caption{\textbf{Qualitative comparison with SOTA approaches}: The above figure shows a sample test image from the TotalText dataset and the end-to-end text spotting performance of DA-TextSpotter compared with others.}
\label{fig:figsotacom}
\end{figure}
Contextual cues from scenes are often used by humans for recognizing text under degraded scene conditions suffering from low-resolution, blurs, distortions, and occluded objects. Despite its practical significance, current scene text recognition (STR) benchmarks~\cite{karatzas2015icdar,ch2020total,liu2019curved} in the literature fail to justify the robustness and effectiveness of state-of-the-art (SOTA) text spotting models under such real-world noisy domains, especially underwater scenes. Capturing clear underwater scene images is practically unfeasible, mostly due to the effect caused by colour scatter in addition to colour cast from varying light attenuation in different wavelengths~\cite{akkaynak2017space, li2019underwater}. The lack of publicly available benchmark led us to develop a real-world \textit{Under-Water Text (UWT)} spotting validation benchmark to further advance the development of end-to-end text spotting systems for multi-domain noisy and complex scene environments.  

Given two visual domains of completely different nature, it often leads to reason that each would require a different architecture. However, multi-domain learning (MDL) was introduced in~\cite{rebuffi2017learning} with the objective to utilize the same computational pipeline to learn incrementally a set of parameters for every added domain, while retaining performance of already learned datasets~\cite{goodfellow2013empirical}. In this work, we explore a new domain-agnostic framework to investigate the ability of a scene text spotting model to generalize to a wider range of domain complexities rather than being specialized for a specific domain. The key intuition behind the utility of MDL to DA-TextSpotter is to improve the model robustness against different variations of noises as demonstrated in ~\cite{roy2018effects,souibgui2022text}. To demonstrate this important property, we propose a transformer-based scene-text spotting pipeline called \textit{Domain Agnostic-TextSpotter (DA-TextSpotter)} which can also generalize to new domains and scenarios (in this case, from natural scene domain to underwater) without requiring any re-training or fine-tuning. 

Intelligent reading systems applied to underwater scenes get affected by reduced visibility and contrast, due to wavelength-dependent absorption and scattering caused by variations of refractive indices associated with temperature, optical turbulence, and salinity microstructures present in lakes and oceans~\cite{hou2012optical}. To mitigate this issue and to significantly improve the capabilities of underwater robotic systems~\cite{wen2023syreanet}, DA-TextSpotter introduces an enhancer unit as a pre-processing step which utilizes the Enhanced-Super-Resolution Generative Adversarial Networks (ESRGAN)~\cite{wang2021real,wang2018esrgan} to enhance and improve the visual quality of the input images coming from either natural scenes or underwater. A Swin-Transformer~\cite{liu2021swin} backbone has been adapted to the proposed framework for visual feature extraction. A significant performance boost is obtained compared to ResNets~\cite{he2016deep}or simple Vision Transformers (ViTs)~\cite{dosovitskiy2020image} owing to the better fine-grained-prediction of text instances with its shifted-window-based attention strategy. The rest of the pipeline follows a single encoder-dual decoder transformer unit, following a similar scheme to TESTR~\cite{zhang2022text}. A further in-depth study with a domain-generalization~\cite{muandet2013domain} setting was also conducted to observe the performance of DA-TextSpotter compared to leading SOTA text spotting approaches. With an overall assessment, DA-TextSpotter outperforms its nearest competitors~\cite{huang2022swintextspotter,fang2022abinet++,zhang2022text} in both text detection and end-to-end recognition tasks for both natural scene and underwater domains.

The key contributions of the paper can be summarized in three folds: 

(1) A novel scene text-spotting framework which is domain-agnostic and achieves superior performance on both natural scene and underwater image benchmarks. 

(2) A simple, efficient and flexible text-spotting pipeline, which introduces a super-resolution-based enhancement unit for dealing with more adverse noisy images, along with a visual feature extraction branch that introduces both local and global attention over patches. 

(3) A new evaluation benchmark for text spotting in underwater scene images which helps to justify the essential findings of this work, along with a new future challenge for the text recognition community.

\section{Related Work}

\noindent
\textbf{Multi-Domain Learning.} Multi-Domain Learning (MDL) was analysed in Joshi \textit{et. al.}~\cite{joshi2012multi} to probe some important research questions on taking the advantage of multiple domain labels to improve learning. Rebuffi {et. al.}~\cite{rebuffi2017learning} then coined it as the task of learning a single visual representation from multiple label spaces of different visual domains, without any forgetting~\cite{li2017learning}. Rosenfeld {et. al.}~\cite{rosenfeld2018incremental} further extended this idea of multi-domain towards a multi-task setting, where different tasks are to be performed on the same domain. Liu {et. al.}~\cite{liu2019compact} trained a network to align and match the feature distributions from multiple spaces for image classification. Motivated by aforementioned works, we use a simple concept of MDL towards scene text spotting across multi-domain complex scenes to gain some model robustness by inherently learning domain-agnostic parameters from diverse noisy domains~\cite{roy2018effects}. \newline  



\noindent
\textbf{Image Super-resolution.} 
The image super-resolution field has evolved since the inception of SRCNN~\cite{dong2014learning} used by super-resolution GAN~\cite{ledig2017photo} models to upsample and enhance low-resolution images and restore them. The ESR-GAN ~\cite{wang2018esrgan, wang2021real} has been one of the landmark super-resolution-based approaches that use a relativistic GAN discriminator~\cite{ronneberger2015u} and a real-world fully synthetic dataset to train their model. This work uses the power of Real-ESRGAN's~\cite{wang2021real} to be used as an enhancement unit in the proposed DA-TextSpotter framework. It becomes really important, especially for text spotting in our proposed UWT benchmark. Previously, Banerjee \textit{et. al.}~\cite{banerjee2022dct} proposed a text detection strategy for underwater scene images, but there has not been any end-to-end text spotting paradigm for such diverse domain.   \newline

\noindent
\textbf{End-to-End Text Spotting.} 
Scene text spotting is considered as joint modelling of text detection and recognition modules, specially challenging in autonomous driving systems~\cite{reddy2020roadtext, garcia2022read, case2011autonomous}. Previous approaches~\cite{wang2011end,liao2017textboxes} addressed the training process in a two-stage manner (detection and recognition) separately and later join them for inference. The first attempt to joint detection and recognition simultaneously were done through RoI operations~\cite{li2017towards,he2018end,liu2018fots} during training, although they had sub-par performance to detect arbitrary-shaped text. Segmentation-based techniques~\cite{lyu2018mask,liao2020mask} based on Mask-RCNN~\cite{he2017mask} became more popular later and achieved decent performance on arbitrary-shaped text instances. However, they were computationally more expensive and PAN++~\cite{wang2021pan++} improved its efficiency by using a faster detector to multiply the segmentation maps with the features to suppress the background as followed in other approaches~\cite{wang2019arbitrary,qin2019towards}. To avoid further post-processing of the mask representation, MANGO~\cite{qiao2021mango} proposed a mask attention module to use more global features and give it more robustness. Other representations for curved texts are parametric Bézier curves used by ABCNet~\cite{liu2020abcnet,liu2021abcnet} or coordinates of bounding-box guided polygon vertices as in TESTR~\cite{zhang2022text}. In this work, we follow a polygonal bounding-box representation for arbitrary-shaped text.  

Modelling linguistic knowledge for end-to-end STR was first proposed in ~\cite{fang2021read} which was later extended to an end-to-end scene text spotting framework in ~\cite{fang2022abinet++} by using  bidirectional and iterative language modelling to spot text regions. Transformers~\cite{vaswani2017attention} became really popular in recent times~\cite{huang2021spatial,xue2023image} in STR tasks due to its global modelling capabilities and better parallelization ability. Later, ViTs~\cite{dosovitskiy2020image} have been used to perform faster and more efficient STR~\cite{atienza2021vision}. Recently, self-supervised ViT pretraining~\cite{souibgui2022text} to learn different kinds of scene degradations has been employed to achieve the best STR performance. For scene text spotting, however, the relationships between different text instances is critical since they might share common background textures and text styles. Current state-of-the-art methods like SwinTextSpotter~\cite{huang2022swintextspotter} and TESTR~\cite{zhang2022text} propose a joint optimization of detection and recognition with different strategies. While SwinTextSpotter uses a Swin transformer\cite{huang2022swintextspotter} backbone coupled with Feature Pyramid Networks (FPNs) to spot smaller text instances, TESTR~\cite{zhang2022text} relies on multi-head deformable attention~\cite{zhu2020deformable} to achieve a similar objective. In this work, we strive towards efficiency by using a tiny variant of Swin feature backbone coupled with deformable attention module for the text spotting unit.  

\begin{figure*}[ht]
\includegraphics[width=\textwidth]{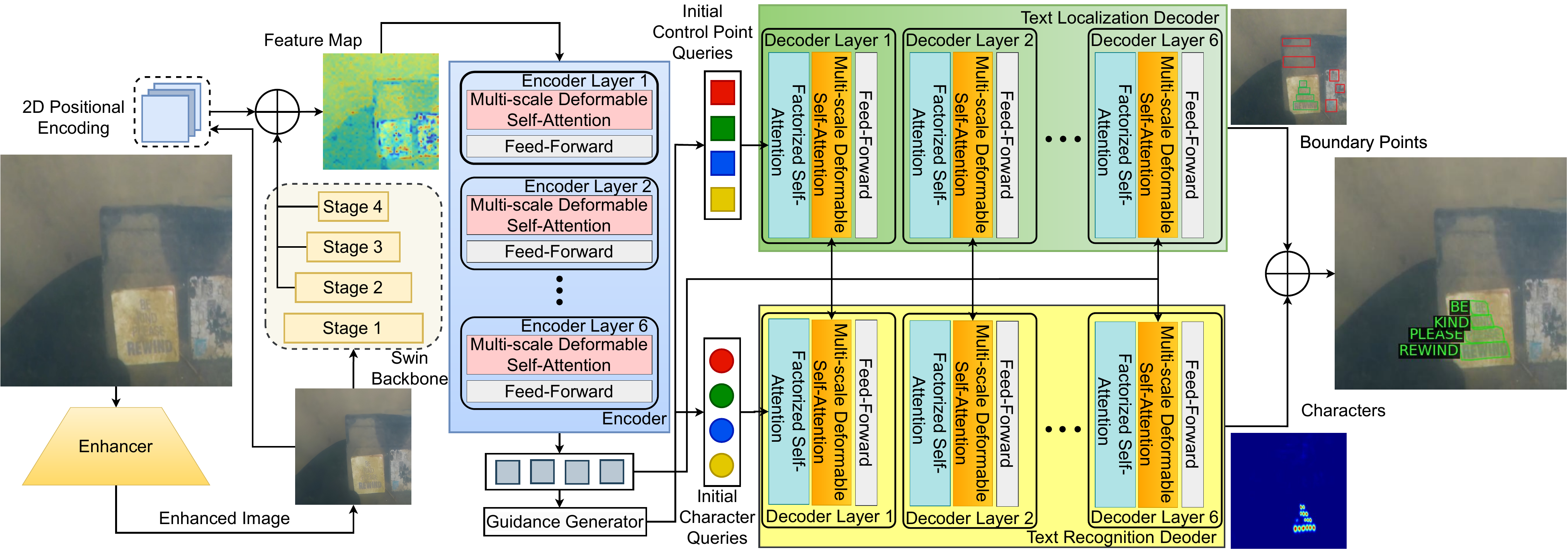}
\caption{\textbf{The overall framework of DA-TextSpotter} consisting of Super-Resolution, Feature Extraction and Text Spotting units sequentially.} \label{fig2}
\end{figure*}

\section{Methodology}

Our method proposes a domain-agnostic and efficient framework called \textit{DA-TextSpotter} for scene text spotting that mainly addresses unified text detection and recognition in noisy domains such as underwater scenes having low-resolution, high contrast, occluded objects and so on. The primary goal of our proposed network is to learn some domain-agnostic feature representation by pre-training on both natural and underwater scene source data. We hereby introduce the architectural pipeline  of our method, describe the training strategies, discuss model efficiency, and finally discuss some key insights.


\subsection{Model Architecture} %
The overall architectural pipeline as illustrated in Fig.\ref{fig2} consists of three major components: (1) a super-resolution unit that helps to pre-process and enhance the input scene images; (2) a feature extraction unit based on a Swin-Transformer~\cite{liu2021swin} backbone to collect scene-context aware information; (3) a text spotting unit which unifies both detection and recognition modules to give us the desired result.

\begin{figure}[ht]
\includegraphics[width=0.5\textwidth]{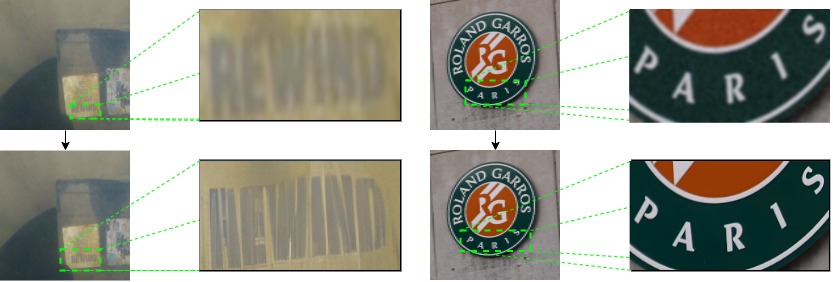}
\caption{Illustration of the \textbf{effectiveness of the Super-Resolution unit} for noise removal on underwater (First case) and natural (second case) scenes. } \label{fig3}
\end{figure}
\noindent
\textbf{Super-resolution Unit.}
The super-resolution unit mainly consists of a pre-trained ESR-GAN~\cite{wang2021real} backbone which has been used for denoising and upsampling the low-resolution and noisy input scene images. The backbone primarily consists of a Residual in Residual Dense Block (RRDB) which combines a multi-level residual network and dense connection without Batch Normalization. In the discriminator, this network uses the relativistic GAN~\cite{jolicoeur2018relativistic} which helps it to generate more realistic images. Using contextual and perceptual losses, the neural network is pushed to the natural or underwater image manifold using a discriminator network trained to differentiate between super-resolved images and original photo-realistic images, whereas adversarial loss pushes the neural network to the natural image manifold. The above details can be quantified in terms of eq.~\ref{eq:1} where $L_G$ denotes the overall loss computed in the generator while $L_1$ and $L_{\text {per}}$ denote the contextual and perceptual losses.

\begin{equation}
    L_G = L_{\text{per}}+\lambda L_G^{R a}+\eta L_1,
\label{eq:1}
\end{equation}

\begin{figure}
\centering
\begin{tabular}{c c c} 
Input images & ResNet-50 & Swin Transformer\\ 
\includegraphics[height=2cm, width= 2cm]{images/0000091_resize.jpg} & \includegraphics[height=2cm, width= 2cm]{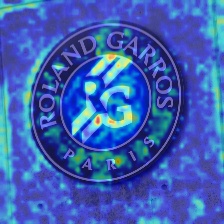}&  \includegraphics[height=2cm, width= 2cm]{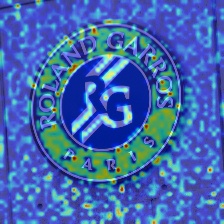} \\
\includegraphics[height=2cm, width= 2cm]{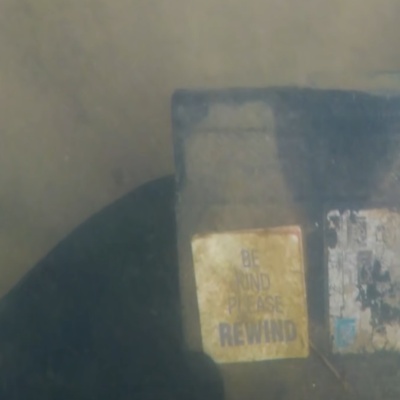} &
\includegraphics[height=2cm, width= 2cm]{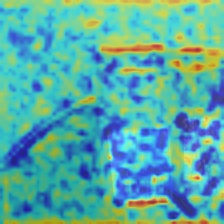}&  \includegraphics[height=2cm, width= 2cm]{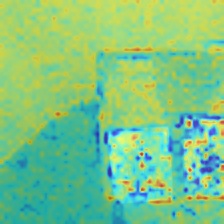} \\ 
\end{tabular} 
\caption{Illustrating the \textbf{advantage of Swin-Transformer} over the ResNet-50 as a backbone generating better-localized representation. }
\label{fig:fig4} 
\end{figure}

\noindent
\textbf{Feature Extraction Unit.} It is hard to connect remote features with vanilla convolutions since they operate locally at fixed sizes (e.g., $3 \times 3$). Text spotting requires modelling the relationship between different texts since scene texts from the same image have strong similarities, such as backgrounds, textures and text styles. For our backbone, we chose to use a small and efficient variant of Swin-Transformer~\cite{liu2021swin} denoted as Swin-tiny. Considering the blanks between words in a line of text, the receptive field should be large enough to help distinguish whether adjacent texts belong to the same line. In Fig.\ref{fig:fig4} we have illustrated how our Swin-tiny backbone manages to generate a better-localized representation over the text than the standard Resnet-50 model. The final output from the last layer of the encoder is then propagated to the next phase.

\noindent
\textbf{Text Spotting Unit.} The text-spotting unit is mainly composed of two primary 
model components: (1) Text Localization Decoder and (2) Text Recognition Decoder. It follows the transformer decoder pipeline from Zhang et. al.~\cite{zhang2022text}. Accordingly, we formulate our problem as a set prediction problem, to predict a set consisting of a set of point-character tuples, for a particular image. We formulate it as $X = \{(S^{(i)}, R^{(i)})\}^K_{i=1}$. Where $i$ is the index of each instances, $S^{(i)} = ({{s^{(i)}_1}, ...., s^{(i)}_N)}$ is the coordinates of $N$ control points, and $R^{(i)} = ({{r^{(i)}_1}, ...., r^{(i)}_N)}$ is the $M$ characters of the text. The location decoder will detect (predict $S^{(i)}$) while the character decoder will recognise (predict $R^{(i)}$) the text in a unified manner.
\newline
\noindent
\textit{Text Localization Decoder.} For the location decoder, we convert the queries to composite queries which predict multiple control points for a single instance. We have such $Q$ queries, each corresponding to a text instance, as $S^{(i)}$. Each query has many sub-queries $s_j$, where $S^{(i)} = ({{s^{(i)}_1}, ...., s^{(i)}_N)}$. Then the initial control points are passed through the location decoder with multiple layers, followed by a classification head that predicts the confidence from the final control points alongside a two-channel regression head generating the normalized coordinates for each. Here the control points are the polygon points starting from the top left corner in clockwise order. 

\noindent
\textit{Text Recognition Decoder.} The character decoder is almost the same as the location decoder, only we change the control points queries with character queries $R^{(i)}$. Both queries, $S^{(i)}$ and $R^{(i)}$ with the same index belong to the same text instance. So, during the prediction, each decoder predicts the control points and characters for the same instance. In the end, a classification head predicts the multiple character classes based on the final character queries. 
 
\subsection{Training Strategies}
The training strategies proposed for the DA-TextSpotter framework have been elucidated in the following two subsections: 

\noindent
\textbf{Domain-agnostic Pretraining.} In this work, we propose a novel training strategy for pretraining which incorporates multiple domain source data from natural scenes and underwater images. The main intuition behind this approach is for the model to learn both domain-independent and domain-specific parameters simultaneously during the pre-training phase which helps them to improve generalization capacity during the fine-tuning phase. Further experimentation has been carried out to effectively justify this strategy in the next section.  

\noindent
\textbf{Learning Objectives.} The overall losses used in this work can be summarised 
under the encoder $\mathcal{L}_{\text {enc }}$ and decoder $\mathcal{L}_{\text {dec }}$ blocks shown in eqns.~\ref{eq:encoder} and ~\ref{eq:decoder} respectively,

\begin{equation}
    \mathcal{L}_{\text {enc }}=\sum_i\left(\lambda_{\text {cls }} \mathcal{L}_{\text {cls }}^{(i)}+\lambda_{\text {coord }} \mathcal{L}_{\text {coord }}^{(i)}+\lambda_{\text {gloU }} \mathcal{L}_{\text {gloU }}^{(i)}\right)
\label{eq:encoder}
\end{equation}

\begin{equation}
    \mathcal{L}_{\text {dec }}=\sum_j\left(\lambda_{\text {cls }} \mathcal{L}_{\text {cls }}^{(j)}+\lambda_{\text {coord }} \mathcal{L}_{\text {coord }}^{(j)}+\lambda_{\text {char }} \mathcal{L}_{\text {char }}^{(j)}\right)
\label{eq:decoder}
\end{equation}

\noindent
Where $\mathcal{L}_{\text {cls }}^{(j)}$ is focal loss for classification of text instances. $\mathcal{L}_{\text {coord }}^{(j)}$ is L-1 loss used  for control point coordinate regression. $ \mathcal{L}_{\text {char }}^{(j)}$ is Cross entropy loss for character classification. $\mathcal{L}_{\text {gloU }}$ is the generalized IoU loss defined in~\cite{rezatofighi2019generalized} for bounding box regression. $\lambda_{\text {cls}}$, $\lambda_{\text {char}}$, $\lambda_{\text {coord }}$, $\lambda_{\text {cls }}$, and $\lambda_{\text {gloU }}$ are the weighting factor for the losses.

\section{Experiments}

\subsection{Datasets and Evaluation}\label{dataset}
\noindent
The UWT dataset~\footnote{The dataset will be made publicly available upon acceptance.} has been proposed in this work to evaluate the performance of the proposed DA-TextSpotter framework. The sample images have been snapped from YouTube videos broadcasting the retrieval of lost items in rivers, seas, and pools. On average, there is just one text instance for each image. All the text instances are in English. Overall, the dataset consists of 250 total images with dimensions of $400 \times 400$ pixels, split into 200 images for training and 50 images for testing. For natural scene benchmarks, CTW1500~\cite{liu2019curved} and TotalText~\cite{ch2020total} has been used for arbitrary-shaped text, while ICDAR15~\cite{karatzas2015icdar} has been used for regular text. The Precision (P), Recall (R) and F1 (F) metrics has been adapted for text detection while end-to-end recognition score use the None (without using lexica) and Full (using lexica) settings. 

\begin{table*}[t]
\centering
\caption{\textbf{Text spotting} performance on Total-Text and CTW1500. ``None'' refers to recognition without lexicon. The ``Full'' lexicon contains all the words in the test set}
\begin{adjustbox}{max width=\textwidth}
\begin{tabular}{@{}lcccccccccc@{}}
\toprule
& \multicolumn{5}{c}{Total-Text}& \multicolumn{5}{c}{CTW1500}\\ \cmidrule(lr){2-6} \cmidrule(lr){7-11}
\multicolumn{1}{c}{\multirow{2}{*}{Methods}}& \multicolumn{3}{c}{Detection}&  \multicolumn{2}{c}{End-to-end} & \multicolumn{3}{c}{Detection}&  \multicolumn{2}{c}{End-to-end} \\ \cmidrule(lr){2-4} \cmidrule(lr){5-6}\cmidrule(lr){7-9}\cmidrule(l){10-11}
 & P & R& F& None& Full& P & R & F& None& Full\\ \hline
ABCNet\cite{liu2020abcnet}&-& -& -& 64.2& 75.7 & -& -& 81.4& 45.2& 74.1\\
Text Perceptron\cite{qiao2020text}&88.8 &81.8 &85.2 &69.7&78.3 & 87.5& 81.9& 84.6 &57.0& -\\
ABCNet v2\cite{liu2021abcnet} &90.2&84.1 &87.0& 70.4 &78.1&  85.6& 83.8& 84.7& 57.5& 77.2 \\
MANGO\cite{qiao2021mango}&-&-&-&72.9&83.6& - & -& -& \textbf{58.9} & 78.7\\
TESTR\cite{zhang2022text}& \textbf{93.4}& 81.4& 86.90& 73.25& 83.3& 89.7& 83.1& 86.3 & 53.3& 79.9\\
Swintextspotter\cite{huang2022swintextspotter}& -& -& 88.0&  \textbf{74.3} & 84.1& -&  -&  88.0&51.8& 77.0\\ \hline
\textbf{DA-TextSpotter}& 91.13& \textbf{86.31} & \textbf{88.66}  & 71.9&\textbf{85.01}& 90.98 &  \textbf{85.53} & 88.17 & 53.96  &82.15\\
\textbf{DA-TextSpotter (w/o UWT)}& 91.21& 85.77 &88.41 &72.56&84.44&\textbf{91.45} & 85.16 &\textbf{88.19} &56.02& \textbf{82.91}\\
\bottomrule
\end{tabular}

\label{tab:my-table4}
\end{adjustbox}
\end{table*}
\subsection{Experiments on Text Spotting}
In this subsection, we highlight the final text spotting results on the different benchmarks.\\

\begin{table}[t]
\centering
\caption{\textbf{Text Spotting} performance on the ICDAR-15 dataset. “S”, “W”, “G” represents recognition with “Strong”, “Weak”, “Generic”, lexica, respectively}
\begin{adjustbox}{max width=0.45\textwidth}
\begin{tabular}{@{}lllllll@{}}
\toprule
\multicolumn{1}{c}{\multirow{2}{*}{Methods}}&\multicolumn{3}{c}{Detection}&& \multicolumn{2}{c}{End-to-end}\\ \cmidrule(lr){2-4} \cmidrule{5-7}
& P& R & F& S& W& G\\ \hline
Unconstrained\cite{qin2019towards}& 89.4& 87.5& 87.5 & 83.4& 79.9& 68.0\\
Text   Perceptron\cite{qiao2020text} & 89.4& 82.5& 87.1& 80.5& 76.6& 65.1  \\ 
ABCNet   v2 \cite{liu2021abcnet}& 90.4& 86.0& 88.1& 82.7& 78.5& 73.0  \\
MANGO \cite{qiao2021mango}& -& -& -& 81.8& 78.9& 67.3  \\
PGNet\cite{wang2021pgnet} & 91.8& 84.8& 88.2& 83.3& 78.3& 63.5  \\
TESTR\cite{zhang2022text}& 90.3&\textbf{89.7}& 90.0& 85.2& 79.4& 73.6  \\
Swintextspotter\cite{huang2022swintextspotter} & -& -& -& 83.9& 77.3& 70.5  \\
Abinet++\cite{fang2022abinet++}& -& -& -&86.1& 81.9& \textbf{77.8 } \\ \hline
\textbf{DA-TextSpotter}& \textbf{92.60} & 88.59  & \textbf{90.55} & 85.92 & 80.44 & 74.16 \\
\textbf{DA-TextSpotter(w/o UWT)}& 91.49 & 89.07  & 90.27 & \textbf{86.78} & \textbf{81.90} & 75.82 \\ \bottomrule
\end{tabular}
\label{tab:my-table6}
\end{adjustbox}
\end{table}

\noindent
\textbf{Arbitrary-shaped Text.} For the irregular text we have used two datasets Total-Text and CTW1500 as described in Section \ref{dataset}. We show our quantitative results in Table \ref{tab:my-table4}.
\noindent
In terms of text detection, we can see our method outperforms TESTR and Swintextspotter in terms of Hmean in the Total-Text dataset. In the case of the CTW1500 dataset, the same surpluses these two methods in terms of all the metrics. 
For text spotting our supervised method outperform all the methods in terms of Full lexicon metrics. In the Total-Text dataset, we outperformed TESTR by almost 1\%. For CTW1500 we outperform TESTR, Swintetextspotter, and Abinet ++ by 3\%, almost 6\%, and almost 2\% respectively.
\noindent

\begin{figure*}[ht]
\centering
\begin{tabular}{c c c c c} 
\includegraphics[height=2cm, width= 2cm]{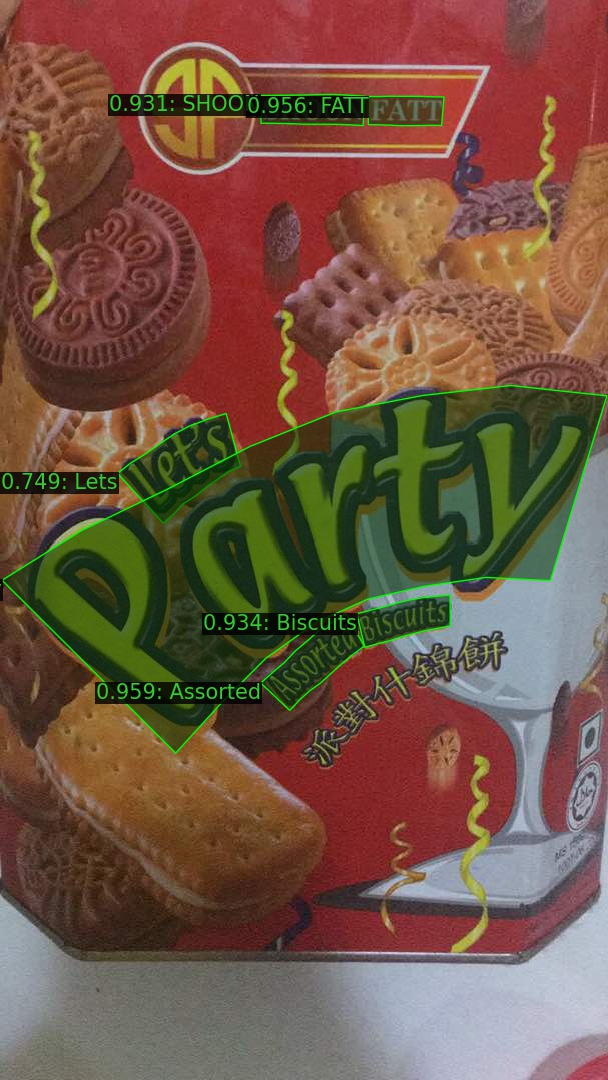}&
\includegraphics[height=2cm, width= 2cm]{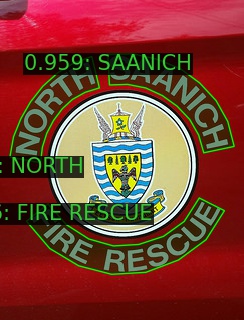}&
\includegraphics[height=2cm, width= 2cm]{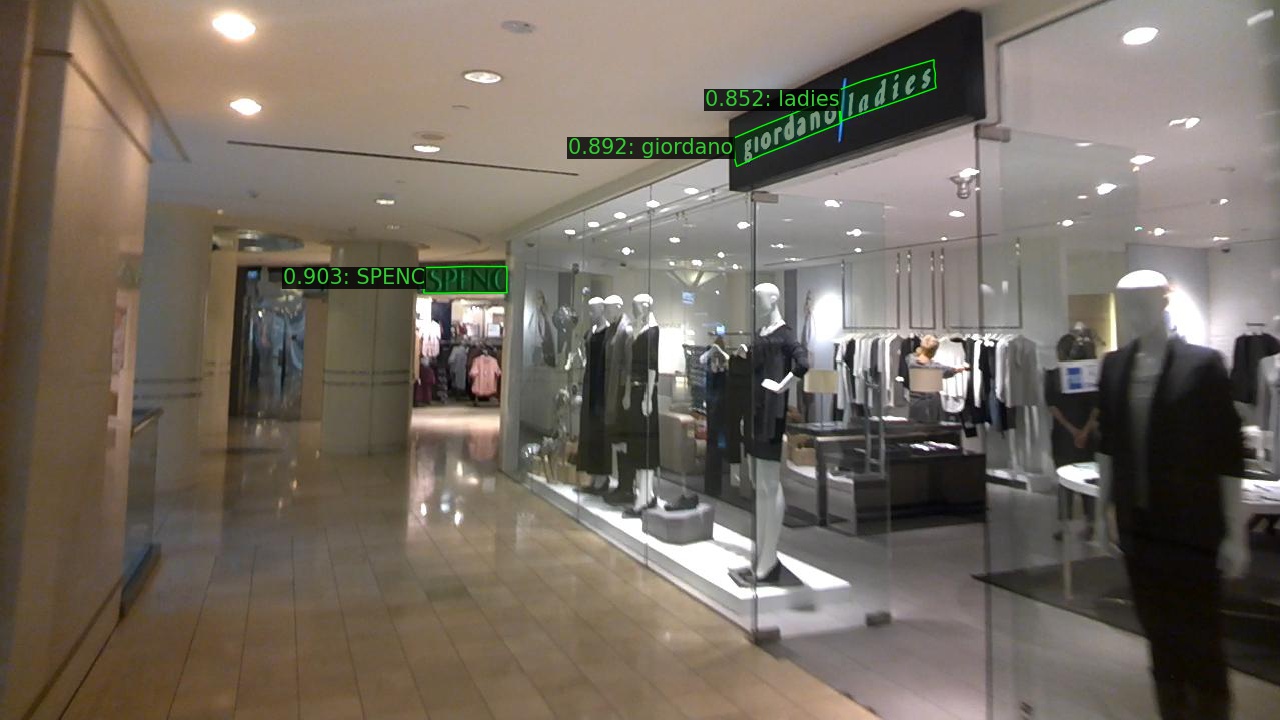}&
\includegraphics[height=2cm, width= 2cm]{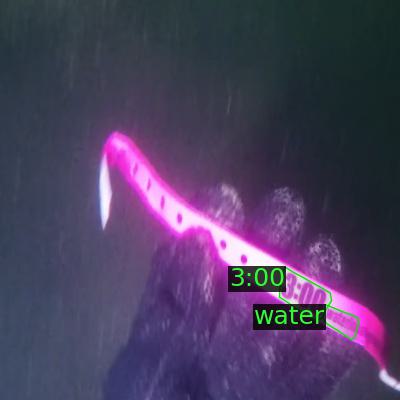}&
\includegraphics[height=2cm, width= 2cm]{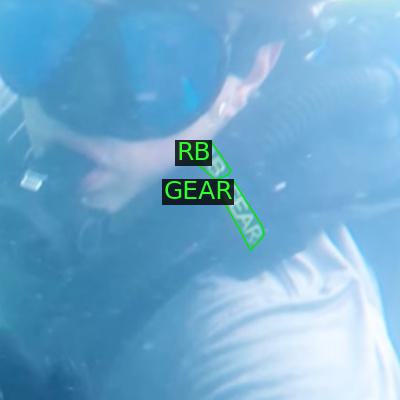}\\
\includegraphics[height=2cm, width= 2cm]{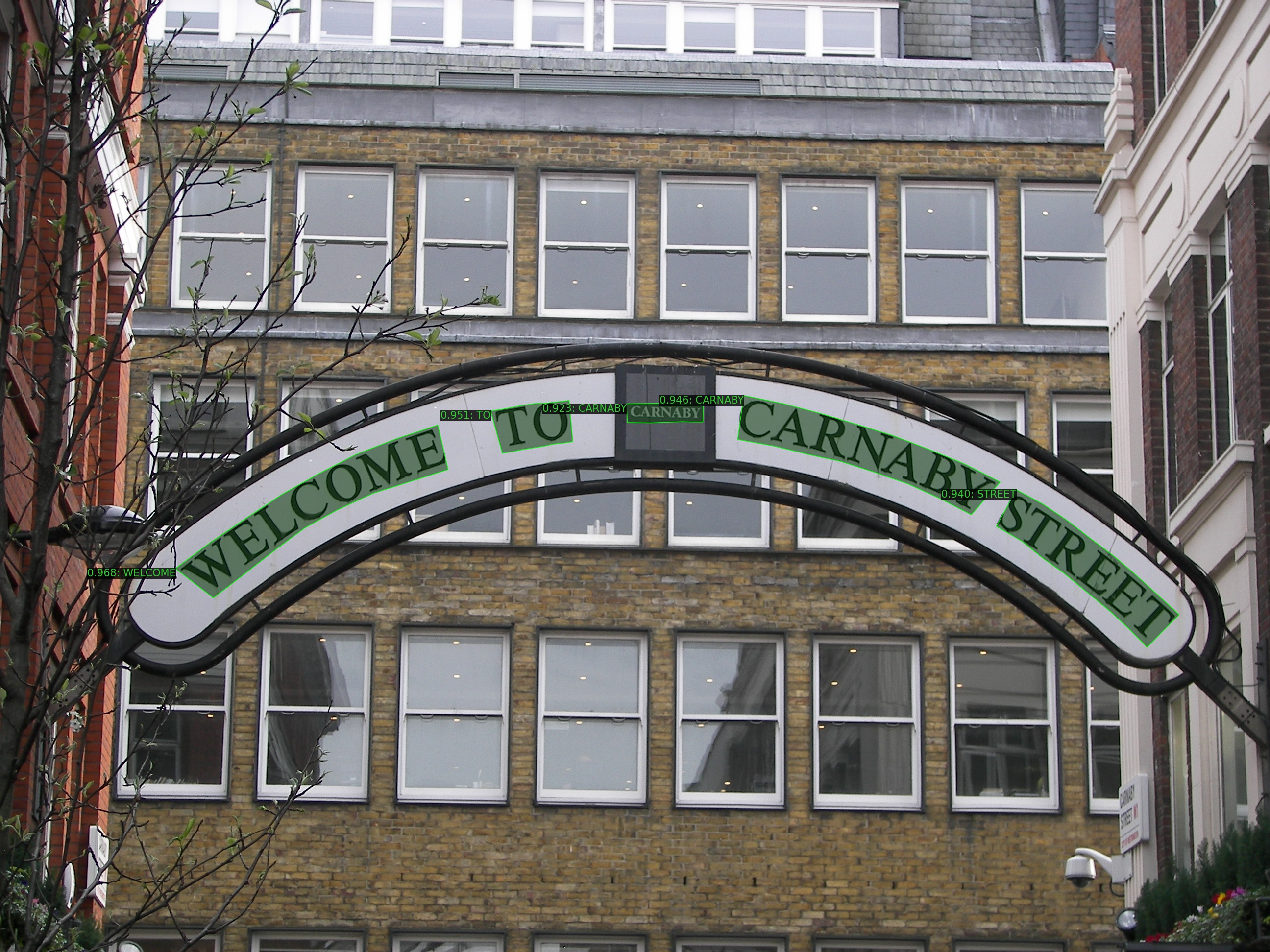}&
\includegraphics[height=2cm, width= 2cm]{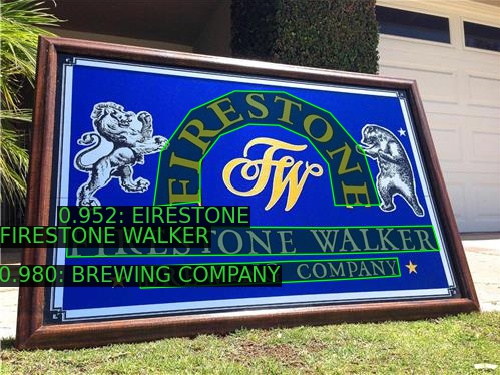}&
\includegraphics[height=2cm, width= 2cm]{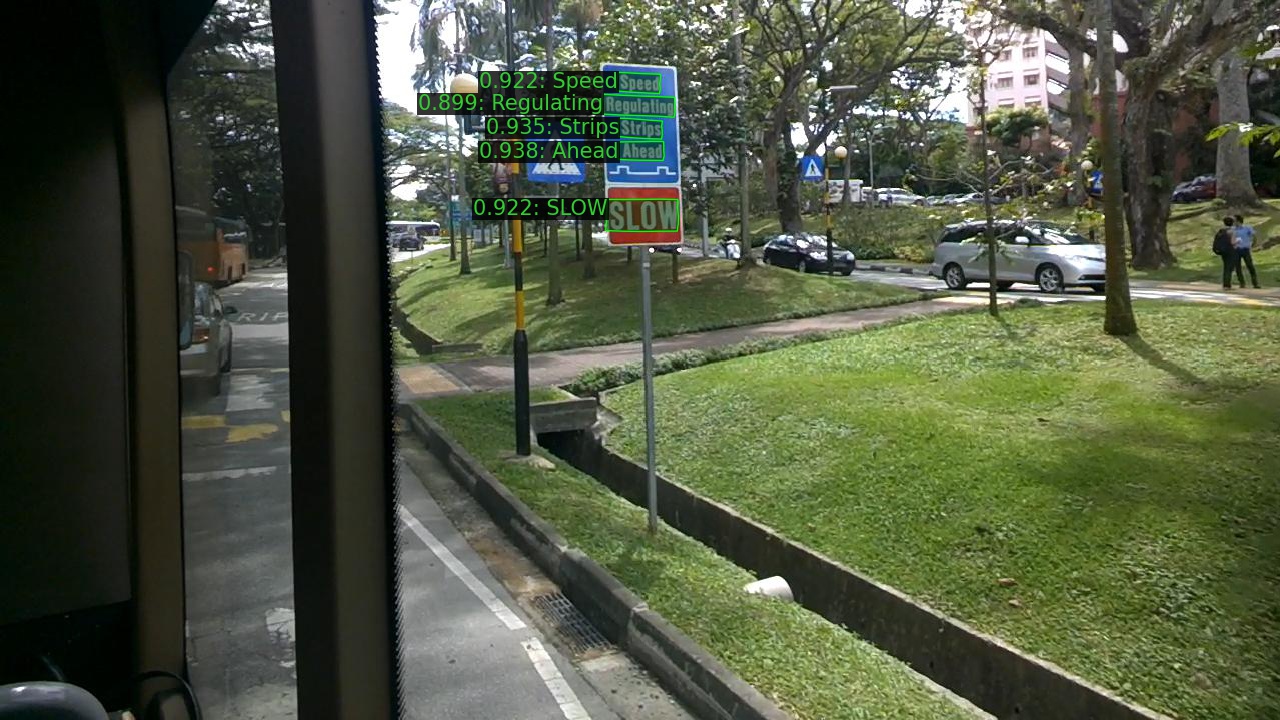}&
\includegraphics[height=2cm, width= 2cm]{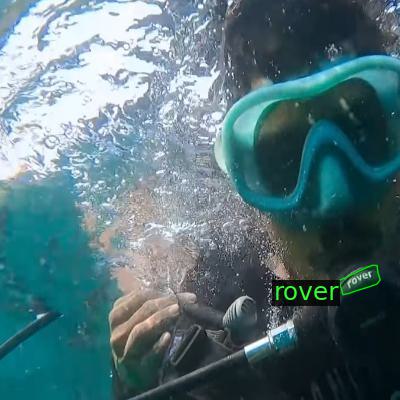}&
\includegraphics[height=2cm, width= 2cm]{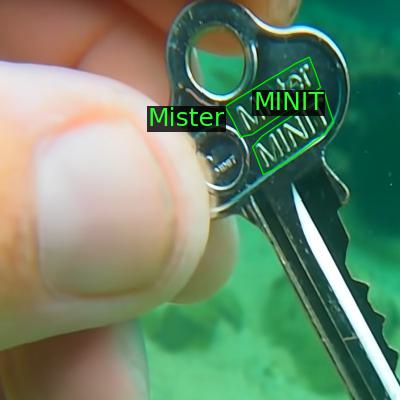}\\
\end{tabular}
\caption{Illustration of our method on different datasets. 1st column from Total-Text, 2nd column from CTW1500, 3rd column from ICDAR15, 4th and 5th columns from UWT. [Use 200\% zoom for better visualization of the qualitative results]}
\label{fig:fig7}
\end{figure*}

\noindent
Qualitative results are shown in Fig~\ref{fig:fig7}. In the first column, we show the results of Total-text and in the second column,  we show the results of CTW1500. Here we can see our method can spot both the regular and irregular text in these datasets. In summary, we can say that the qualitative and quantitative results represent the effectiveness of our model. \newline
\noindent
\textbf{Regular Text.} ICDAR15 has been used as a regular text benchmark. The performance of DA-TextSpotter compared to SOTA approaches is shown in Table ~\ref{tab:my-table6}. For Text detection, we achieve a 2\% gain in precision which lead to achieve slightly gain in the F-measure compared to TESTR. In the text spotting task, our method gives the best result in the ``Strong'' type outperforming TESTR, Swintextspotter, and ABINet++ by almost 1\%, 2\%, and 0.5\% respectively. In Fig \ref{fig:fig7}, 3rd column shows how our method performs on ICDAR15.\\
\noindent
\textbf{UWT.} In UWT we compare it with SOTA methods and our previous method. We can see almost 5\%, 3\%, and 4\% improvement compared with Banerjee \textit{et. al.} in terms of detection Precision, Recall, and F-measure respectively. Table \ref{tab:my-table7} shows how we outperform the SOTA methods in terms of all the metrices. We also show some qualitative results of our method in Fig \ref{fig:fig7} (last two columns).

\begin{table}[t]
\centering
\caption{\textbf{Text Spotting} performance of the proposed and the SOTA systems on UWT.}
\begin{adjustbox}{max width=0.65\textwidth}
\begin{tabular}{@{}lccccccc@{}}
\toprule
\multicolumn{1}{c}{\multirow{2}{*}{Methods}}&\multicolumn{3}{c}{Detection}& &\multicolumn{2}{c}{End-to-end} &\multicolumn{1}{c}{\multirow{2}{*}{\#Params}}\\ \cmidrule(lr){2-4} \cmidrule{6-7} 
& P& R & F& & \multicolumn{2}{c}{None}\\ \hline
Banerjee et. al.~\cite{banerjee2022dct} & 90.25& 45.37& 60.38 &&&-&- \\ 
TESTR\cite{zhang2022text}& 92.24& 33.86& 49.54&& &29.63& \textbf{55M }  \\
Swintextspotter~\cite{huang2022swintextspotter} & 83.21 & 34.49 & 48.77 &&&29.08& 150M \\
\hline
DA-TextSpotter& \textbf{95.65} & \textbf{48.73}  & \textbf{64.57} &&&\textbf{64.15} & 60M \\ \bottomrule
\end{tabular}
\end{adjustbox}
\label{tab:my-table7}
\end{table}

\begin{table}[ht]
\centering
\caption{\textbf{Domain Generalization} settings showing the effectiveness of adding UWT}
\begin{adjustbox}{max width=0.48\textwidth}
\begin{tabular}{@{}cccccccc@{}}
\toprule
\multirow{2}{*}{Dataset}& \multirow{2}{*}{Method}& \multicolumn{3}{c}{Detection} & End-to-end \\  \cmidrule(lr){3-5} \cmidrule(lr){6-6} 
& & P& R & F& None \\ \hline
\multirow{2}{*}{ICDAR2015}& DA-TextSpotter(w/o UWT) & 83.41&84.74& 84.07& 52.20\\
& DA-TextSpotter & \textbf{91.08}& \textbf{84.59}& \textbf{87.72}& \textbf{84.02}\\ \hline
\multirow{2}{*}{Total-Text} & DA-TextSpotter(w/o UWT)& 88.42& 67.25& 76.40&59.77\\
&DA-TextSpotter &\textbf{91.76}& \textbf{76.42}&\textbf{83.39}&\textbf{64.67}\\ \hline
\multirow{2}{*}{CTW1500}& DA-TextSpotter(w/o UWT)& 39.19&36.56& 37.71&0\\
&DA-TextSpotter & \textbf{41.69}&\textbf{44.35}&\textbf{42.98}&\textbf{0}\\
\bottomrule
\end{tabular}
\end{adjustbox}
\label{tab:my-table8}
\end{table}
\subsection{Domain Generalization}
Normally it is seen that if we train our model with one type of image and after that if we train it with noisy images it losses its performance drastically its called the catastrophic forgetting phenomenon of deep learning neural network. To address this phenomenon we perform zero-shot experiments. First, we train our model with Synthtext and Icdar17 dataset namely Pre-train Scene Text. We evaluate it on the SOTA datasets, Total-text, CTW1500, and ICDAR15. After that, we again train it with Underwater images. As it consists of more noisy images. We expected it will degrade the performance of the model. But in Table \ref{tab:my-table8} it improves the result. In that case, the End-to-End result of CTW1500 is 0 because it is text-line based annotated, and in the pretraining, our model only learns word-level annotation.

\begin{table}[ht]
\centering
\caption{Ablation for different feature extraction backbones.  Where ``P'' stands for Precision, ``R'' stands for Recall, ``F'' stands for F-measure respectively, and ``None'' refers no lexicon has been used.}
\begin{adjustbox}{max width=0.48\textwidth}
\begin{tabular}{@{}lcccc@{}}
\toprule
\multirow{2}{*}{Methods} & \multicolumn{3}{c}{Detection}& \multicolumn{1}{c}{End-to-end}\\ \cmidrule(lr){2-4} \cmidrule(lr){5-5} 
& P & R & F &  None \\ \hline
Resnet-50 \cite{he2016deep} & 88.87& 76.47& 82.20&    60.06  \\
ViT-T \cite{dosovitskiy2020image}& 90.17& 72.90  & 80.62  & 59.40\\ 
\textbf{Swin-T} \cite{liu2021swin} & \textbf{93.59} &\textbf{75.20} &\textbf{83.40} & \textbf{67.06} \\ 
\hline
\label{tab:table1}
\end{tabular}
\end{adjustbox}
\end{table}

\begin{table}[ht]
\centering
\caption{Results for different enhancement methods on UWT.  Where ``P'' stands for Precision, ``R'' stands for Recall, ``F'' stands for F-measure respectively, and ``None'' refers no lexicon has been used.}
\begin{adjustbox}{max width=0.48\textwidth}
\begin{tabular}{@{}lcccc@{}}

\toprule
\multicolumn{1}{c}{\multirow{2}{*}{Methods}}&\multicolumn{3}{c}{Detection}& \multicolumn{1}{c}{End-to-end}\\ \cmidrule(lr){2-4} \cmidrule{5-5} 
& P& R & F &  None\\ \hline
Without Enhancement & 68.06 & 31.01 & 42.61 &16.09 \\
SRGAN \cite{ledig2017photo}& 85.54& 40.76& 55.21 &56.32 \\ 
ESRGAN\cite{wang2018esrgan}& 90.20& 42.56 & 57.83 &  58.32 \\ \hline
Real-ESRGAN\cite{wang2021real} & \textbf{95.65} & \textbf{48.73} & \textbf{64.57} &\textbf{64.15} \\
\bottomrule
\label{tab:table3}
\end{tabular}
\end{adjustbox}
\end{table}

\subsection{Ablation Studies}
To understand the significance of the different components of the DA-TextSpotter framework, we conduct ablation studies demonstrated as follows: \newline
\noindent
\textbf{Effectiveness of Swin Transformer.} The usage of Swin-Tiny~\cite{liu2021swin} feature extraction backbone in the DA-TextSpotter framework shows a significant performance gain over ViT-Tiny~\cite{dosovitskiy2020image} and Resnet-50~\cite{he2016deep} backbones in both detection and recognition metrics.  Almost a 4\% change in detection precision and a 1\% change in the detection F-measure over Resnet-50 is observed. Further, we get almost a massive 7\% improvement in the End-to-End recognition results when it is run on the Total-Text~\cite{ch2020total} dataset as shown in Table~\ref{tab:table1}. This justifies that the window-based local attention computed with Swin-T can be really effective for specially text regions with smaller local boundaries. Fig.~\ref{fig:fig4} illustrates how hierarchical feature maps help for feature extraction, especially for smaller objects depicted in the underwater images.

\begin{table}
\centering
\caption{\textbf{Representation Quality.} Fine-tuning performance of a pre-trained model using frozen backbones. Also, 'None' refers to no lexicon has been used.}
\begin{adjustbox}{max width=0.45\textwidth}
\begin{tabular}{lccccccc}
\toprule
\multicolumn{3}{c}{Architectural blocks}&\multicolumn{3}{c}{Detection}&\multicolumn{2}{c}{End-to-End}\\ \cmidrule(lr){1-3} \cmidrule(lr){4-6} \cmidrule(lr){7-8}
\multirow{2}{*}{\shortstack{Image\\ Encoder} } & \multirow{2}{*}{\shortstack{Location\\ Decoder}} & \multirow{2}{*}{\shortstack{Character\\ Decoder}} &  \multirow{2}{*}{P} & \multirow{2}{*}{R} & \multirow{2}{*}{F} &&\multirow{2}{*}{None}\\
& & & & &\\\hline
\multicolumn{1}{c}{\color{red}\xmark}&\multicolumn{1}{c}{\color{red}\xmark}& \multicolumn{1}{c}{\color{teal}\cmark}&92.09&84.64 &88.21&&70.76\\
\multicolumn{1}{c}{\color{teal}\cmark}&\multicolumn{1}{c}{\color{red}\xmark}&\multicolumn{1}{c}{\color{red}\xmark}&92.05&80.49&85.88&&68.9\\
\multicolumn{1}{c}{\color{red}\xmark}&\multicolumn{1}{c}{\color{teal}\cmark}& \multicolumn{1}{c}{\color{red}\xmark}&90.22&85.82&87.69&&69.75\\
\multicolumn{1}{c}{\color{red}\xmark}&\multicolumn{1}{c}{\color{red}\xmark}& \multicolumn{1}{c}{\color{teal}\cmark}&\textbf{93.29}&82.88&87.7&&72.54\\
\multicolumn{1}{c}{\color{teal}\cmark}&\multicolumn{1}{c}{\color{red}\xmark}& \multicolumn{1}{c}{\color{teal}\cmark}&93.23&80.08&86.57&&\textbf{73.39}\\
\multicolumn{1}{c}{\color{teal}\cmark}&\multicolumn{1}{c}{\color{teal}\cmark}&\multicolumn{1}{c}{\color{red}\xmark}&92.88&84.28&88.37&&70.74\\
\multicolumn{1}{c}{\color{teal}\cmark}& \multicolumn{1}{c}{\color{teal}\cmark}& \multicolumn{1}{c}{\color{teal}\cmark}&91.21&\textbf{85.77}&\textbf{88.41}&&72.56\\
\bottomrule
\end{tabular}
\end{adjustbox}
\label{tab:table2}
\end{table}

\noindent
\textbf{Effectiveness of Super-resolution Unit.} For justifying the importance of the Super-resolution unit, we conducted an ablation on our proposed UWT benchmark which contains the most complex low-resolution images. Upsampling the images with the different Super-resolution GAN variants which include SR-GAN~\cite{ledig2017photo}, ESR-GAN~\cite{wang2018esrgan}, and Real-ESRGAN~\cite{wang2021real}. The performance of Real-ESRGAN is extremely high compared to other approaches because of the pre-trained OutdoorSceneTraining~\cite{wang2018recovering} dataset, giving an almost 7\% improvement over both detection F-measure and end-to-end metrics over second-best ESRGAN as shown in Table~\ref{tab:table3}. Also enhancing the images shows a substantial gain of \textbf{22\%} and \textbf{48\%} for detection and end-to-end metrics respectively. Fig.~\ref{fig3} illustrates how this enhancement unit can help to read under complexities like underwater scenes where text is illegible for even human vision. 

\noindent
\textbf{Representation Quality.} To evaluate the representation quality of the features learned during pre-training and the need for the encoder and dual decoder blocks in the DA-TextSpotter pipeline we did an exhaustive evaluation by freezing and unfreezing them. We performed all the possible combinations of frozen and unfrozen decoder and encoder to show the representation quality. All the results are illustrated in Table~\ref{tab:table2} performed on the Total-Text dataset in the fine-tuning configuration. The consistent performances on both detection and recognition tasks for all the different configurations  show the model's robustness.

\section{Conclusion and Future Work}
This work demonstrates the capability of a scene text spotting model to improve its robustness and generalization capability when trained across multiple domains, in this case natural scenes and underwater. The super-resolution unit introduced in the DA-TextSpotter framework could boost performance substantially in more adverse and noisy underwater images. DA-TextSpotter also brings efficiency while using a tiny Swin-transformer backbone for extracting visual cues, both at the local and global level of hierarchies. Finally, multi-domain learning could help a model to learn domain-agnostic parameters, without forgetting already learned domain-specific parameters. The future challenge includes going deeper into domain-incremental settings and using a faster attention module to improve text spotting performance.    


{\small
\bibliographystyle{ieee_fullname}
\bibliography{egbib}
}
\end{document}


\title{Diving into the Depths of Spotting Text in Multi-Domain Noisy Scenes - Supplementary Material}

\author{Alloy Das\\
Computer Vision and Pattern Recognition Unit \\
Indian Statistical Institute \\
Kolkata, India\\
{\tt\small alloyuit@gmail.com}
\and
Sanket Biswas\\
Computer Vision Center\\
Universitat Aut`onoma de Barcelona\\
Barcelona, Spain\\
{\tt\small sbiswas@cvc.uab.es}
\and
Umapada Pal\\
Computer Vision and Pattern Recognition Unit \\
Indian Statistical Institute \\
Kolkata, India\\
{\tt\small umapada@isical.ac.in}
\and
Josep Llad´os\\
Computer Vision Center\\
Universitat Aut`onoma de Barcelona\\
Barcelona, Spain\\
{\tt\small josep@cvc.uab.es}
}




\maketitle



\paragraph{This is the supplementary version for the main paper. Please read the main version for the model formulation, its performance analysis and key ablation experiments. All our pre-trained models and code will be made available at \url{https://github.com/xxx/DA-TextSpotter}.}


\section{Datasets} 

The ground truth for text detection and recognition has been created by manual annotation. A representative selection of examples from the UWT has been shown in Fig.\ref{fig:fig5}, which illustrates different complexities in terms of both the background and the foreground text instances. \newline
\noindent
For state-of-the-art comparison, the following benchmarks have been selected for experimental validation:\\
\noindent
\textbf{ICDAR 2015}\cite{karatzas2015icdar} is the official dataset for ICDAR 2015 robust reading competition for regular text spotting. It has 1000 training and 500 testing images.
\\
\noindent
\textbf{Total-text} \cite{ch2020total} is a well-known arbitrary-shaped text spotting benchmark with 1255 training and 300 testing images. The text instances in this benchmark are at the word level.
\\
\noindent
\textbf{CTW1500}\cite{liu2019curved} is another arbitrary-shaped text spotting benchmark consisting of 1000 training and 500 testing images. The text instances in this benchmark are at the multi-word level.

\section{Evaluation Metrics} The standard metrics Recall (R), Precision (P), and F-measure (F) have been used for evaluating text detection performance. While the end-to-end performance evaluations of text recognition use the None (without using lexica) and Full (using lexica) settings. Furthermore, the Strong (S), Weak (W), and Generic (G) lexica settings have been used to measure recognition performance. For all images, the S, W, and G settings correspond to a small lexicon with 100 words, a lexicon with all the words in the test set, and a lexicon with 90,000 words, respectively. By varying the lexicon sizes, we test the recognition performance under different settings. The proposed system has been evaluated against the state-of-the-art using standard evaluation protocols used in these datasets, which are based on calculating IoU between the predicted and ground-truth polygons. 

\section{Implementation Details}
The ESRGAN has been trained with DIV2K~\cite{agustsson2017ntire}, Flickr2K~\cite{timofte2017ntire}, and OutdoorSceneTraining~\cite{wang2018recovering} benchmarks. The High Resolution (HR) patch size for training is 265. The model has been trained for 1000K iterations with a learning rate of $2 \times 10^{-4}$ using the Adam optimizer \cite{kingma2014adam}. The hyper-parameters for the deformable transformer~\cite{zhu2020deformable} have been kept similar to the original work, with the number of heads = 8 and using 4 sampling points for deformable attention. The number of encoder and decoder layers are initialized to 6.

\noindent
\textbf{Data augmentation.} During pre-training, we augment the data with a random resize with the shorter edge from 480 to 896 and the longest edge is kept within 1600. An instance-aware random crop has also been used.

\noindent
\textbf{Pre-training.} The model is pre-trained with synthtext150k~\cite{liu2020abcnet}, MLT 2017~\cite{nayef2017icdar2017} and TotalText~\cite{ch2020total} for 4000K iteration with a learning rate of $2 \times 10^{-5}$ and is decayed at 3000-Kth iteration by a factor of 0.1. AdamW~\cite{loshchilov2017decoupled} has been used as an optimizer, with $\beta_1 = 0.9$, $\beta_2= 0.999$ and a weight decay of $ 10^{-5}$. We use $Q = 100$ composite queries. The number of control points is 20 and the maximum text length is set to 25. The whole pre-training process has been implemented on one RTX 3080 Ti GPU with a batch size of 1 for an overall 16 days.  

\noindent
\textbf{Fine-tuning.} After pre-training, we fine-tune our model with the specific dataset. For the Total-Text, ICDAR2015, and UWT datasets, we train our model with 200K iterations. For the CTW1500 dataset, we fine-tune our model over 2000K iterations with a maximum length of text 100. As it is annotated sentence level it needs more iterations to train.

\begin{figure}[h]
  \begin{center}
    \begin{tabular}{c c} 

\includegraphics[height=2.2cm, width= 2.2cm]{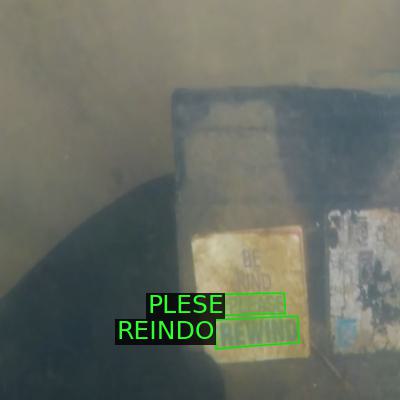}  & \includegraphics[height=2.2cm, width= 2.2cm]{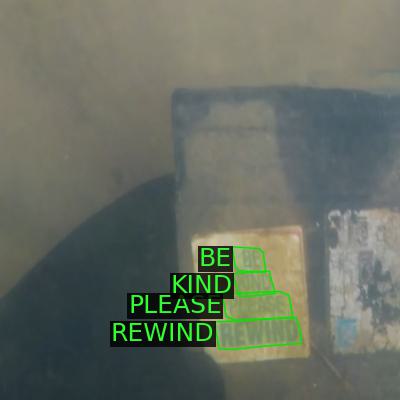} \\  \includegraphics[height=2.2cm, width= 2.2cm]{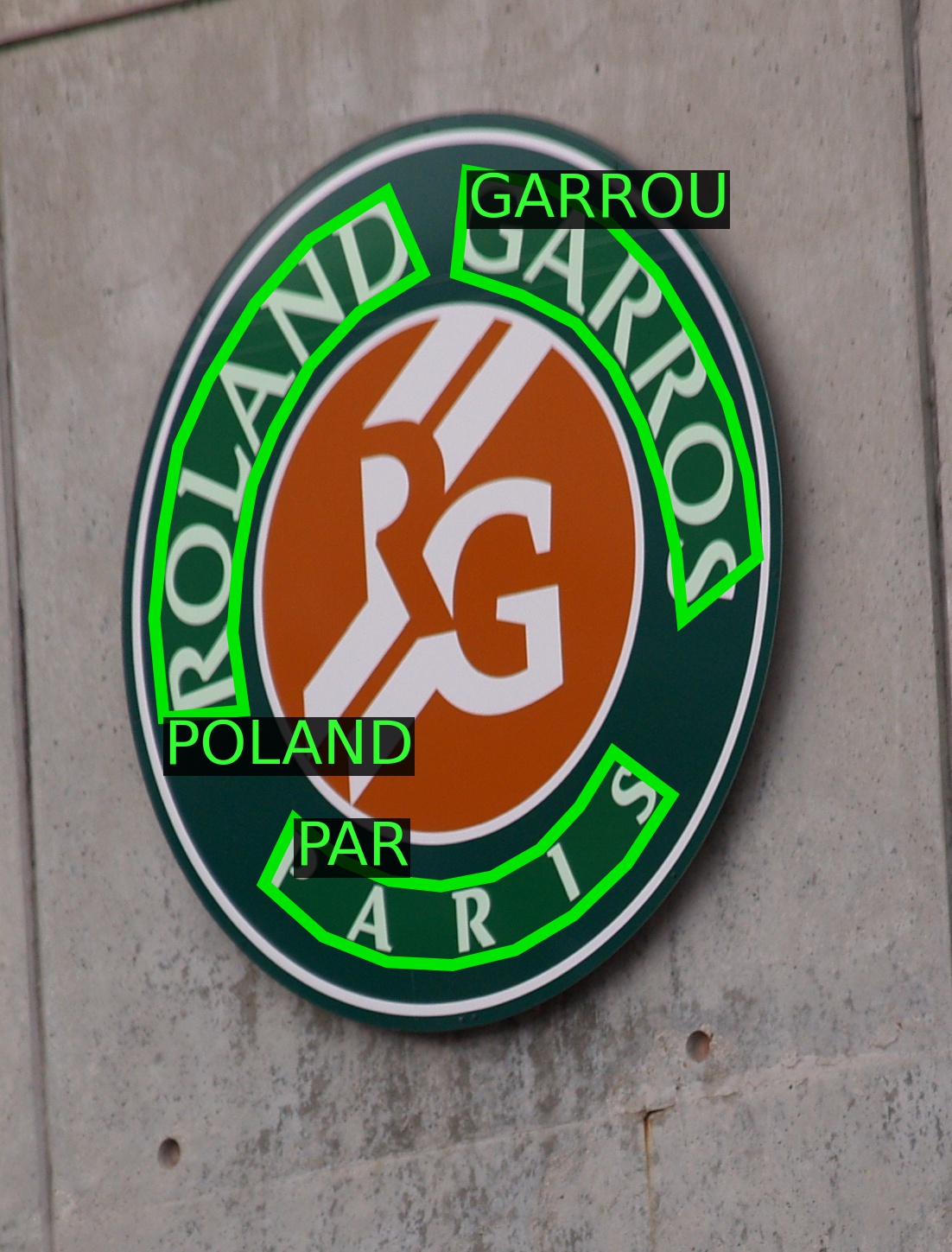} &
\includegraphics[height=2.2cm, width= 2.2cm]{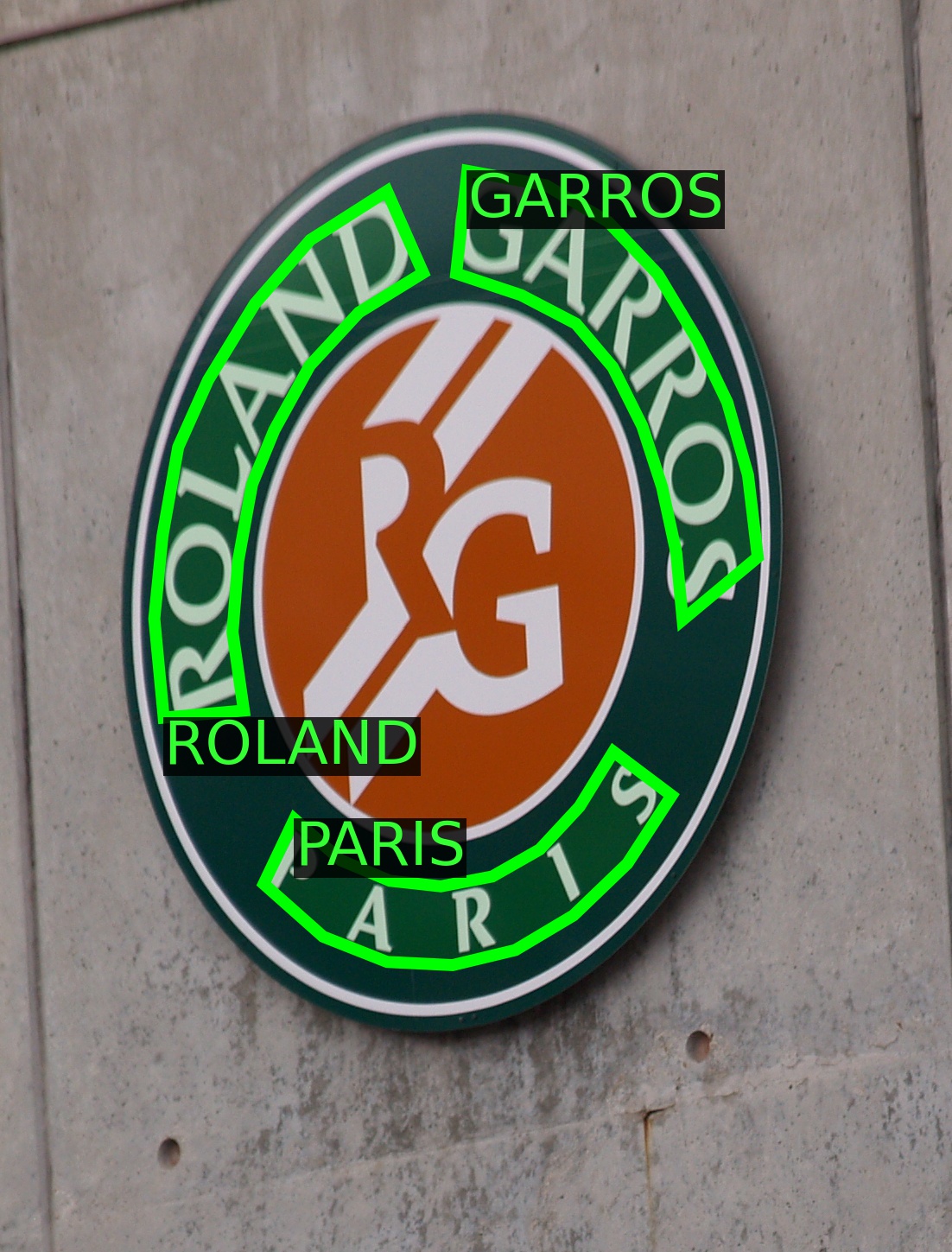} \\ 
\end{tabular} 
  \end{center}
  \caption{ An illustration of the performance of a representative state-of-the-art method and the proposed method for an example of an underwater scene image and one natural scene image. first and third are the Performance of the state-of-the-art approach of Zhang et al.\cite{zhang2022text}. Second and Forth are the Performance of the proposed system.
 }
 \label{fig:fig1}
\end{figure}

\begin{figure}[t]
\centering
\begin{tabular}{c c c c} 
  \includegraphics[height=1.5cm, width= 1.5cm]{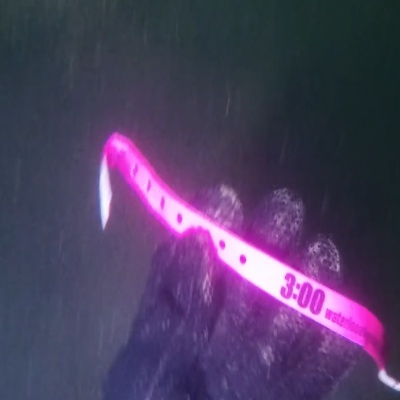}&
  \includegraphics[height=1.5cm, width= 1.5cm]{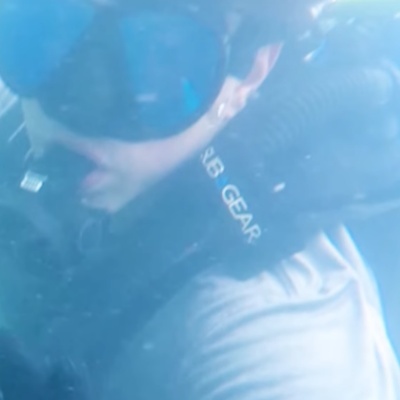}&
  \includegraphics[height=1.5cm, width= 1.5cm]{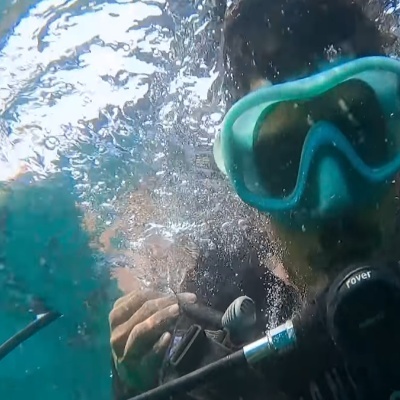}&
  \includegraphics[height=1.5cm, width= 1.5cm]{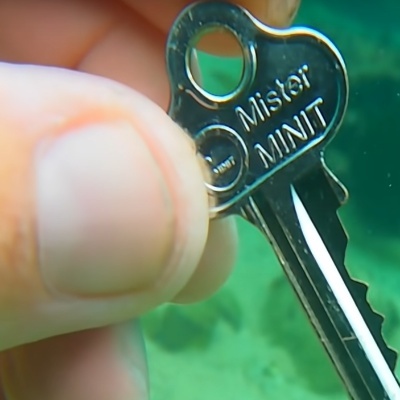}\\
\end{tabular}
\caption{Representative sample instances from the proposed UWT benchmark}
\label{fig:fig5}
\end{figure}

\begin{figure}
\centering
\begin{tabular}{c c c c} 
  \includegraphics[height=1.5cm, width= 1.5cm]{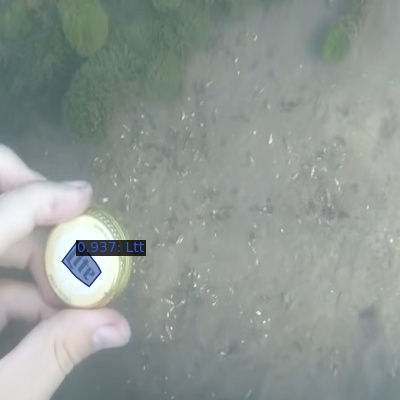}&
  \includegraphics[height=1.5cm, width= 1.5cm]{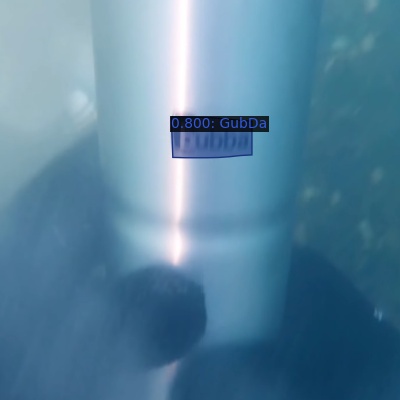}&
  \includegraphics[height=1.5cm, width= 1.5cm]{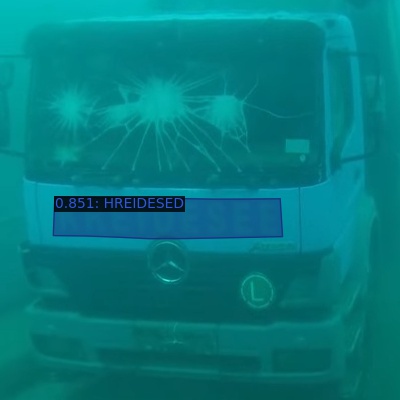}&
  \includegraphics[height=1.5cm, width= 1.5cm]{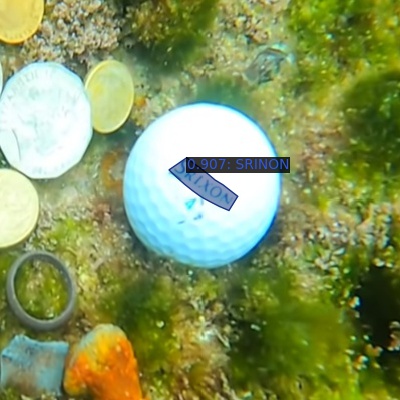}\\
\end{tabular}
\caption{Some failure cases of our method on DA-TextSpotters. Zoom in for better visualization.}
\label{fig:fig9}
\end{figure}

\begin{figure}[t]
    \centering
    \includegraphics[width = 0.48\textwidth]{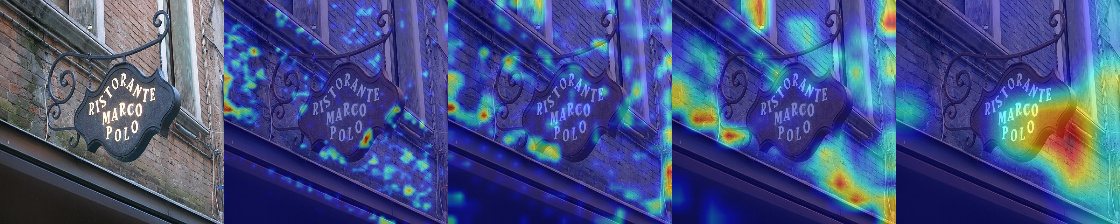}
    \caption{\textbf{Visualization of attention maps for Resnet-50 feature backbone}. (L) to (R) shows attention maps starting from the first layer.}
    \vspace{-0.302cm}
\end{figure}
\begin{figure}[t]
    \centering
    \includegraphics[width = 0.48\textwidth]{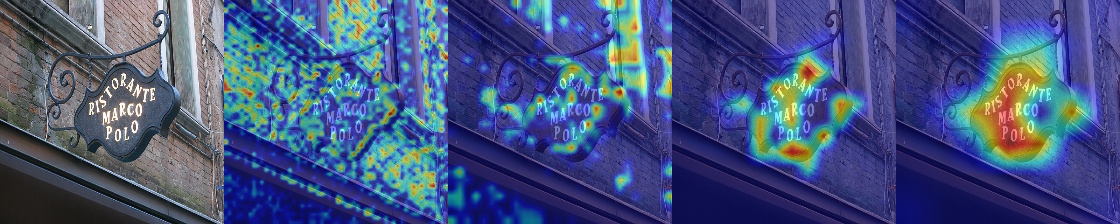}
    \caption{\textbf{Visualization of attention maps for Swin-Tiny feature backbone}. (L) to (R) shows attention maps from the first layer.}
\end{figure}

\begin{figure*}[ht]
\centering
\begin{tabular}{c c c } 
\includegraphics[height=4cm, width= 4cm]{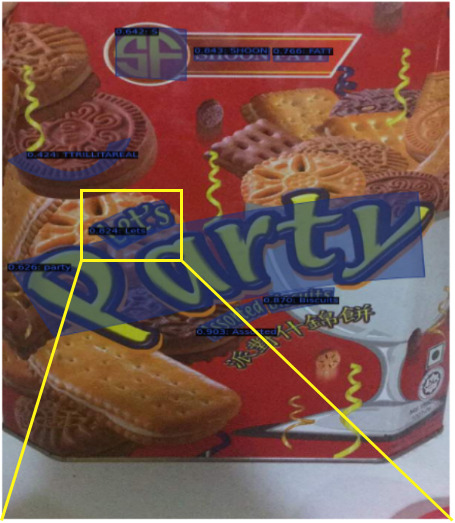}&
\includegraphics[height=4cm, width= 4cm]{images/abcnet v2_.jpg}&
\includegraphics[height=4cm, width= 4cm]{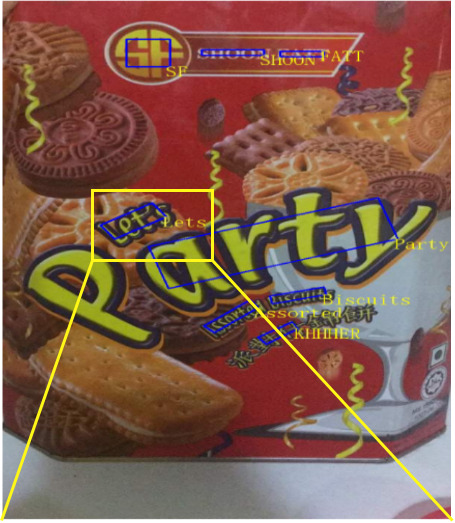}\\
\includegraphics[height=4cm, width= 4cm]{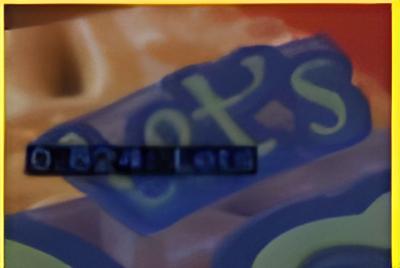}&
\includegraphics[height=4cm, width= 4cm]{images/abc v2_crop.jpeg}&
\includegraphics[height=4cm, width= 4cm]{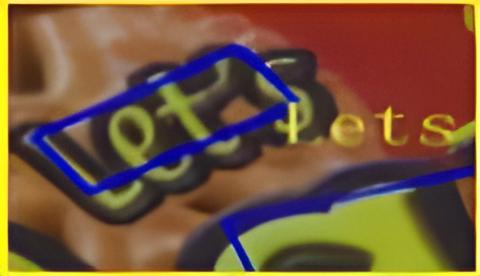}\\
(a) ABCNet &(b) ABCNetv2 &(c) MANGO \\
\includegraphics[height=4cm, width= 4cm]{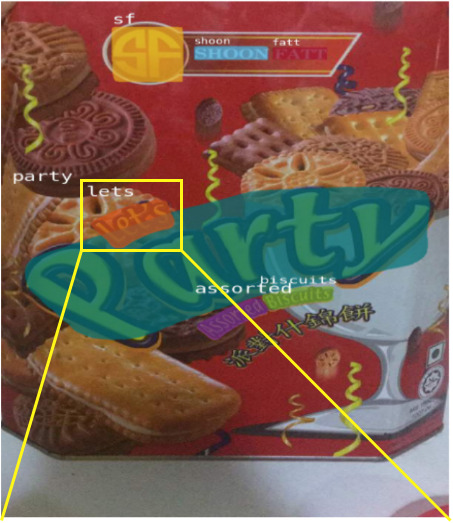}&
\includegraphics[height=4cm, width= 4cm]{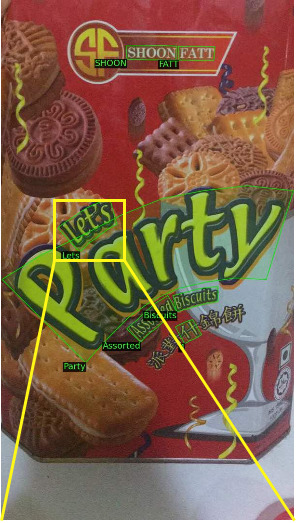}&
\includegraphics[height=4cm, width= 4cm]{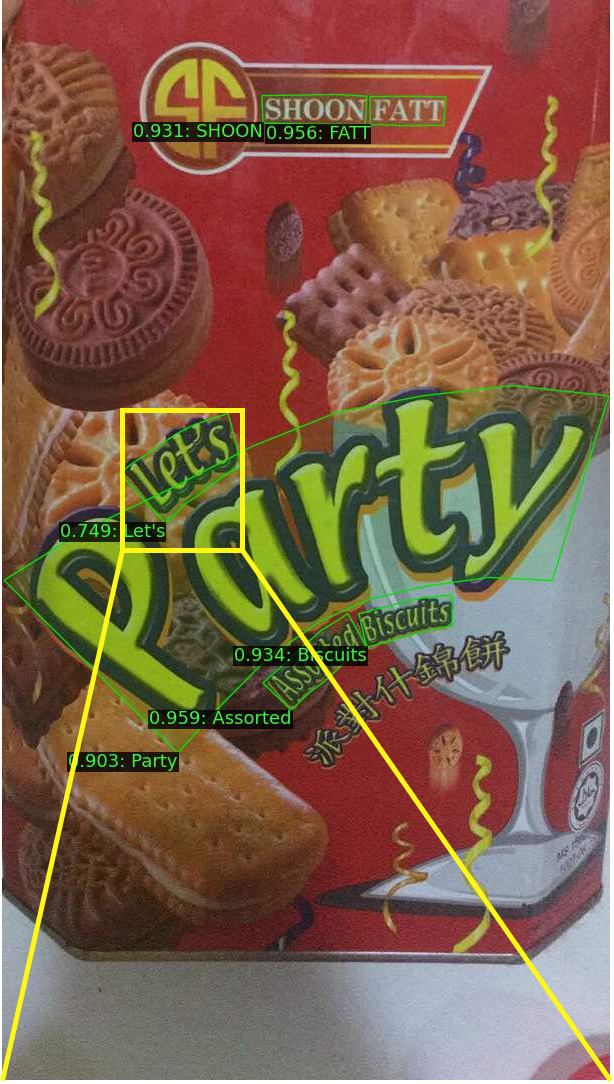}\\
\includegraphics[height=4cm, width= 4cm]{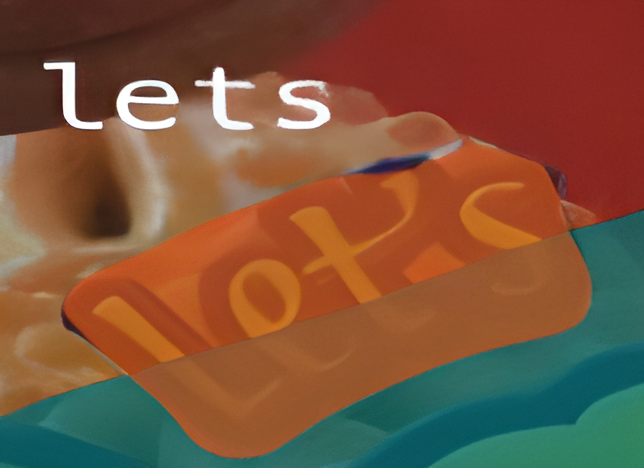}&
\includegraphics[height=4cm, width= 4cm]{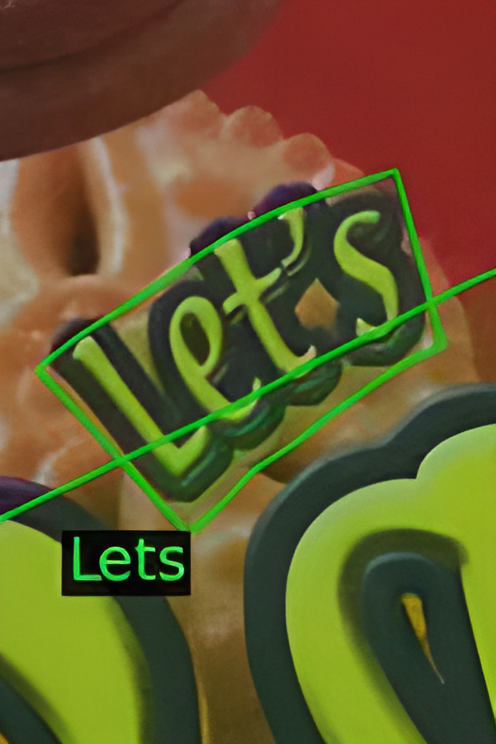}&
\includegraphics[height=4cm, width= 4cm]{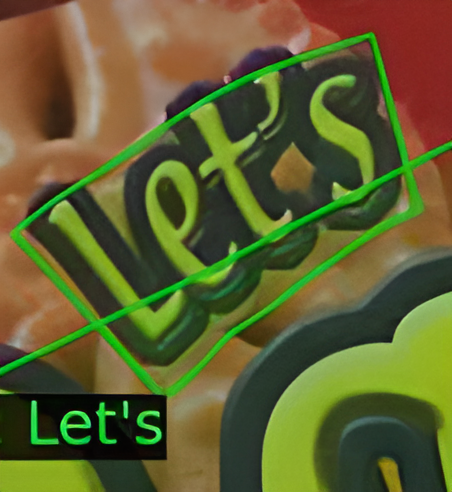}\\
(d) SwinTextSpotter &(e) TESTR & (f) \textbf{DG-TextSpotter}\\
\end{tabular}
\caption{\textbf{Qualitative comparison with SOTA approaches}: The above figure shows a sample test image from the TotalText dataset and shows the end-to-end text spotting performance of DA-TextSpotter when compared to existing SOTA methods. We show a more fine-grained comparison with all the competitors.}
\label{fig:figsotacom}
\end{figure*}

\paragraph{Evaluation Metrics} The standard metrics Recall (R), Precision (P), and F-measure (F) have been used for evaluating text detection performance. While the end-to-end performance evaluations of text recognition use the None (without using lexica) and Full (using lexica) settings. Furthermore, the Strong (S), Weak (W), and Generic (G) lexica settings have been used to measure recognition performance. For all images, the S, W, and G settings correspond to a small lexicon with 100 words, a lexicon with all the words in the test set, and a lexicon with 90,000 words, respectively. By varying the lexicon sizes, we test the recognition performance under different settings. The proposed system has been evaluated against the state-of-the-art using standard evaluation protocols used in these datasets, which are based on calculating IoU between the predicted and ground-truth polygons.

\section{Further Discussions and Insights}

\paragraph{More Experiments on Encoder and Decoder } We have performed some experiments on the possible combinations of the number of decoder and encoder layers to show the trade-off between the choice of the number of encoder/decoder layers with detection and end-to-end performance. All the results are illustrated in Table \ref{tab:my-table10}. The experiments are performed on the Total-Text dataset in the fine-tuning configuration. We observe that with the uniform number of layers (6) for the encoder and decoder blocks, we achieve the most optimal performance for both detection and end-to-end spotting.

\begin{table}[h]
\centering
\caption{Experiments on the number of Encoder and Decoder layers.}
\begin{tabular}{@{}cccccc@{}}
\toprule
 \multicolumn{2}{c}{No of Layers} & \multicolumn{3}{c}{Detection} & End-to-end \\\cmidrule(lr){1-2} \cmidrule(lr){3-5} \cmidrule(lr){6-6}
Encoder&Decoder& P & R & F &   None \\ \hline
4 & 6  & \textbf{90.85}     & 74.03  & 81.58  &  59.80  \\
6 & 4  & 90.17     & 72.90  & 80.62  & 59.40  \\
6 & 6  & 88.87    & \textbf{76.47} & \textbf{82.20} & \textbf{60.06 } \\
\bottomrule
\end{tabular}

\label{tab:my-table10}
\end{table}

\paragraph{How does DA-TextSpotter fare qualitatively when compared to other approaches?} As shown in Fig.~\ref{fig:figsotacom}, we show an example test case from TotalText~\cite{ch2020total} and analyse how our method fares compared to other competitors. We realise that DG-TextSpotter is the only method which could spot the apostrophe (') special character in the word "Let's". Also it spots the words in a case-sensitive manner, contrary to SwinTextSpotter~\cite{huang2022swintextspotter} which is our closest competitor. Also, the high confidence scores of the spotted text predictions in our model when compared to other approaches like ABCNet series~\cite{liu2020abcnet, liu2021abcnet} or MANGO~\cite{qiao2021mango} are also very noticeable. This also justifies the superiority of the quantitative performance of our model when compared to others.     

\paragraph{To what extent does Swin-Tiny feature backbone benefit DA-TextSpotter?} We show a comparison of attention maps visualized in different layers for ResNet-50 backbone~\cite{he2016deep} when compared to the Swin-Tiny~\cite{liu2021swin} variant. The Swin attention better captures the global interactions between the different objects to localize the text region which helps them to get a substantial gain over ResNet-50 which primarily captures more local spatial relationships with the attention. The hierarchical shifted-window-based mechanism is highly useful to control the attention on top of the text region boundaries to have a more complete understanding using better contextual cues.    

\begin{figure*}[]
\centering
\begin{tabular}{c c c c} 
\includegraphics[height=2.5cm, width= 2.5cm]{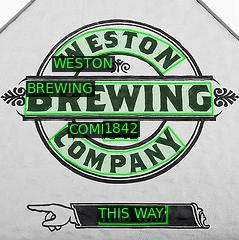}&
\includegraphics[height=2.5cm, width= 2.5cm]{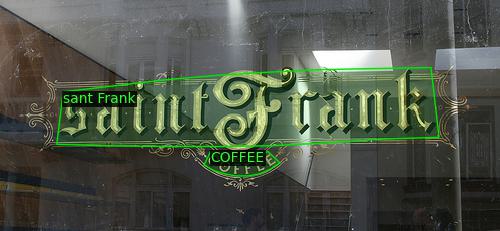}&
\includegraphics[height=2.5cm, width= 2.5cm]{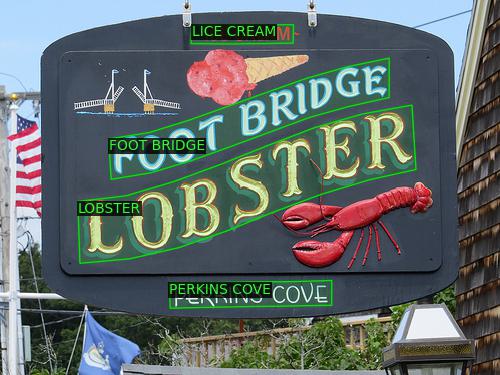}&
\includegraphics[height=2.5cm, width= 2.5cm]{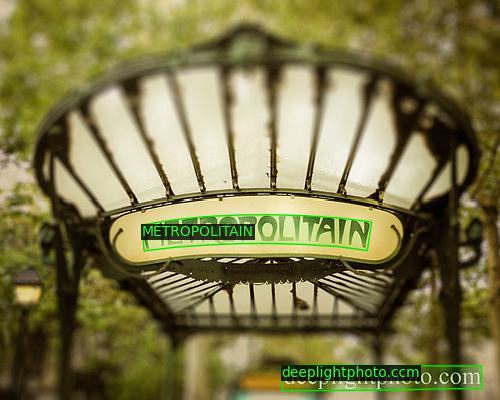}\\
(a)&(b)&(c)&(d)\\
\includegraphics[height=2.5cm, width= 2.5cm]{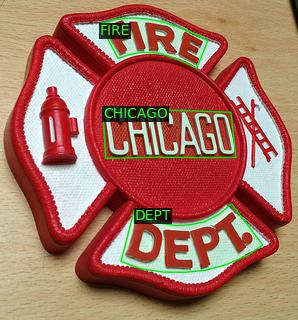}&
\includegraphics[height=2.5cm, width= 2.5cm]{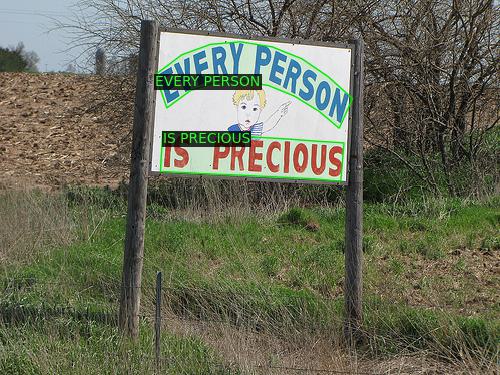}&
\includegraphics[height=2.5cm, width= 2.5cm]{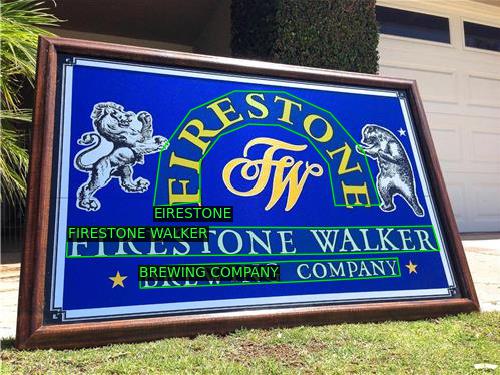}&
\includegraphics[height=2.5cm, width= 2.5cm]{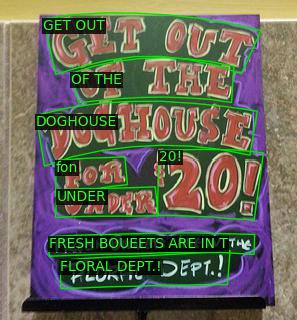}\\
(e)&(f)&(g)&(h)\\
\includegraphics[height=2.5cm, width= 2.5cm]{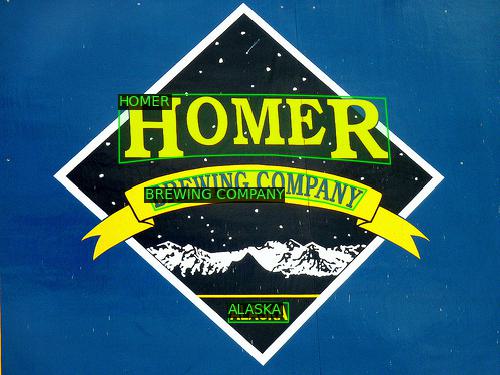}&
\includegraphics[height=2.5cm, width= 2.5cm]{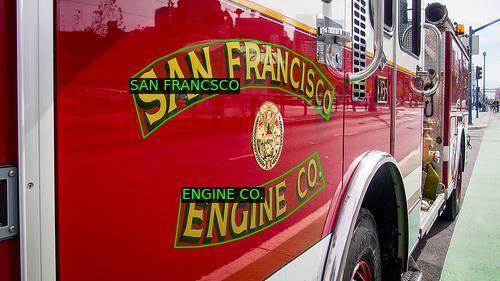}&
\includegraphics[height=2.5cm, width= 2.5cm]{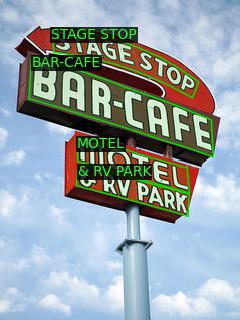}&
\includegraphics[height=2.5cm, width= 2.5cm]{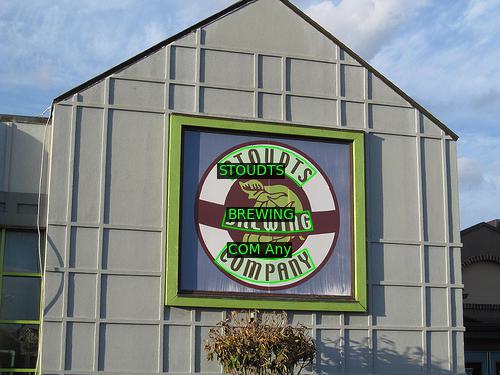}\\
(i)&(j)&(k)&(l)\\
\end{tabular}
\caption{\textbf{Performance Analysis on CTW1500.} Some qualitative test cases of our method on images from CTW1500.}
\label{fig:figsupctw}
\end{figure*}

\paragraph{Qualitative Performance Analysis on CTW1500} As shown in Fig.~\ref{fig:figsupctw}, CTW1500 is a challenging benchmark when it comes to spotting arbitrary-shaped text at the sentence level. Examples (f) and (h) show how DG-TextSpotter is able to spot multiple words together correctly even though it's primarily pre-trained on word-level datasets.   

\begin{figure*}[]
\centering
\begin{tabular}{cccc} 
\includegraphics[height=2.5cm, width= 2.5cm]{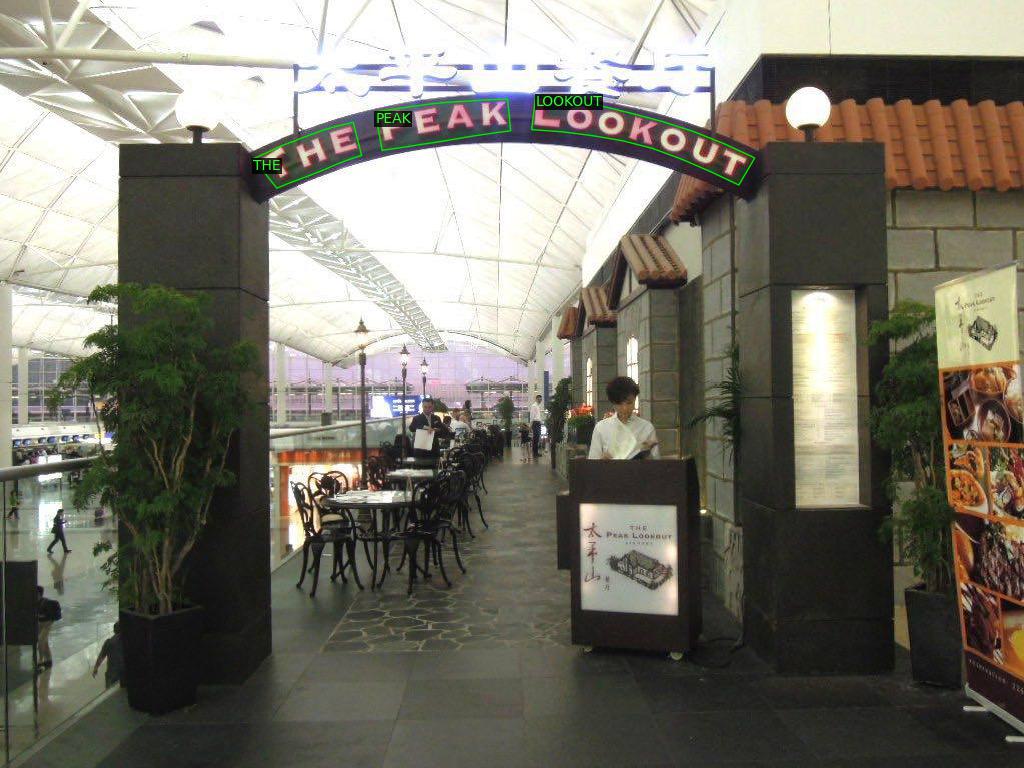}&
\includegraphics[height=2.5cm, width= 2.5cm]{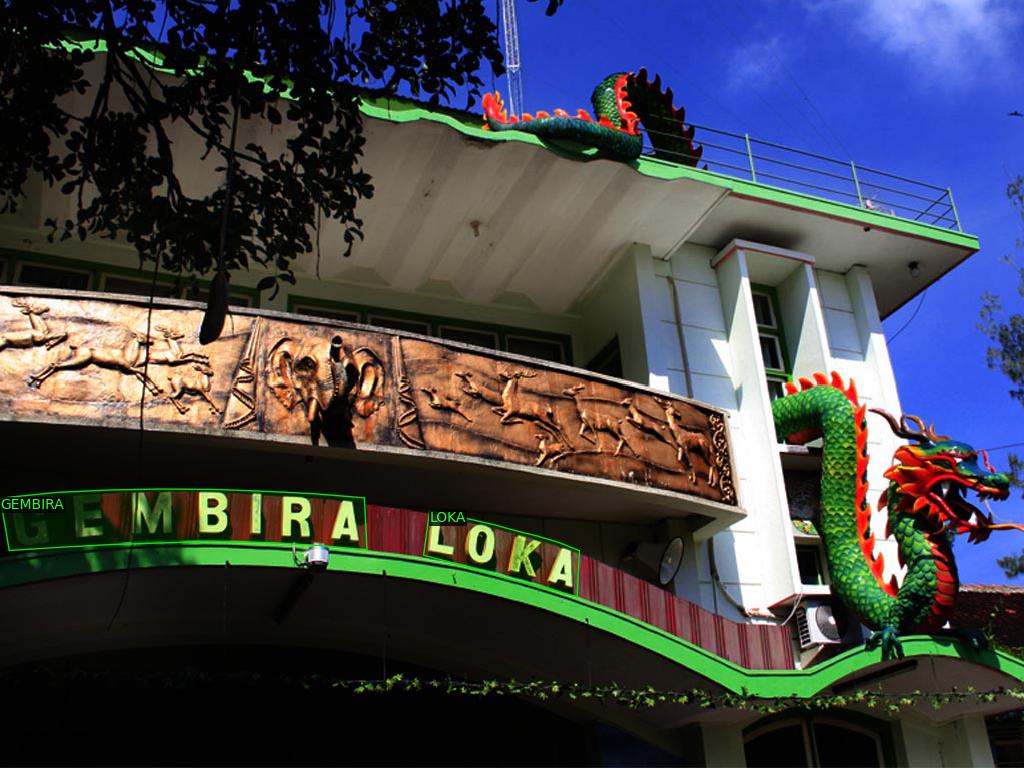}&
\includegraphics[height=2.5cm, width= 2.5cm]{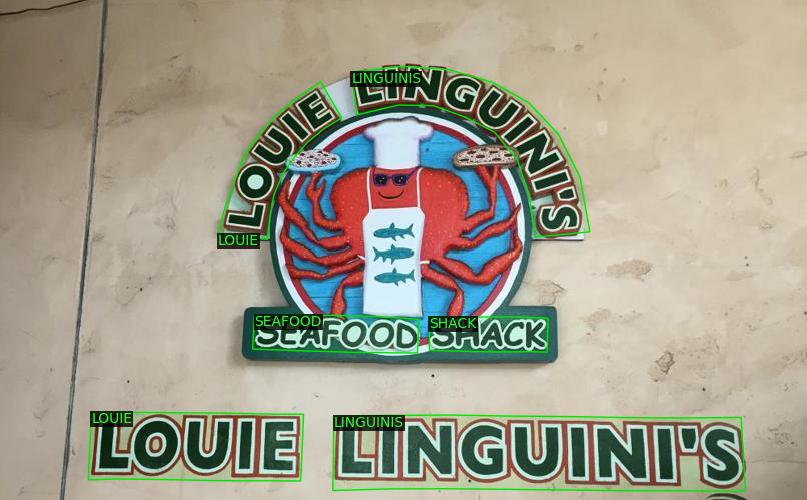}&
\includegraphics[height=2.5cm, width= 2.5cm]{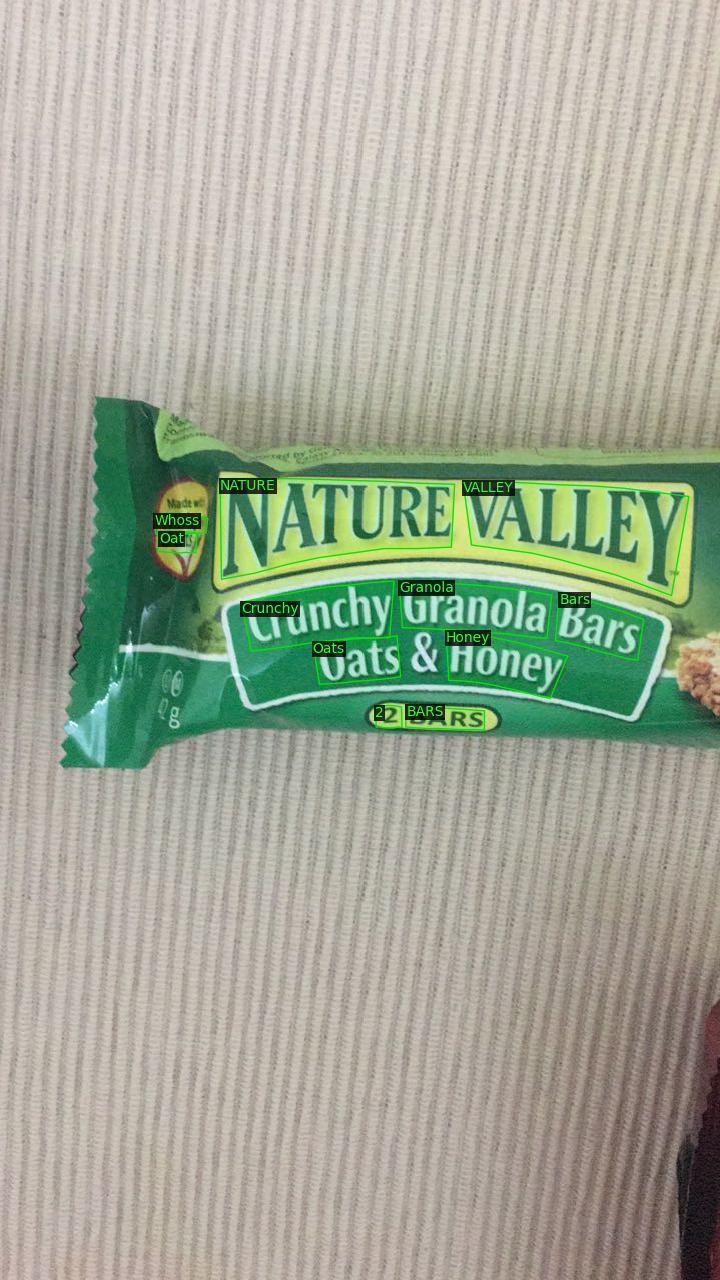}\\
(a)&(b)&(c)&(d)\\
\includegraphics[height=2.5cm, width= 2.5cm]{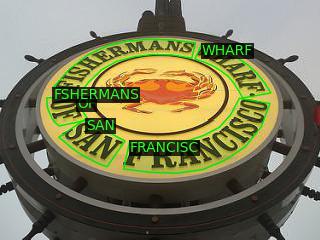}&
\includegraphics[height=2.5cm, width= 2.5cm]{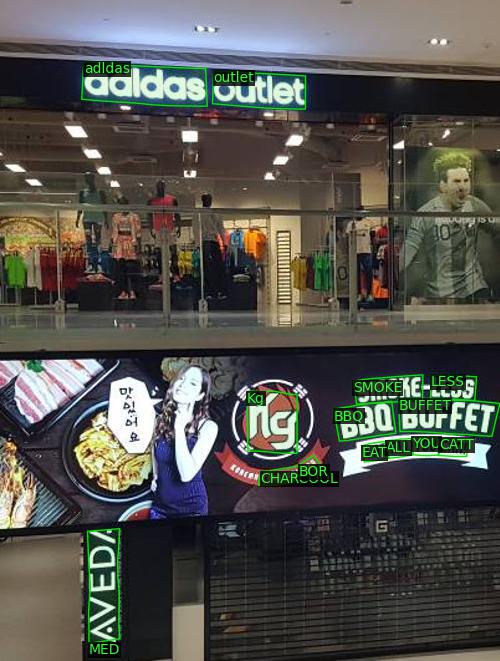}&
\includegraphics[height=2.5cm, width= 2.5cm]{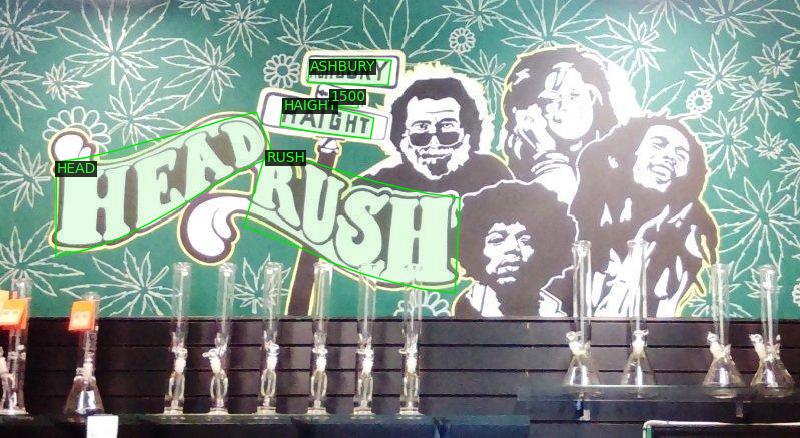}&
\includegraphics[height=2.5cm, width= 2.5cm]{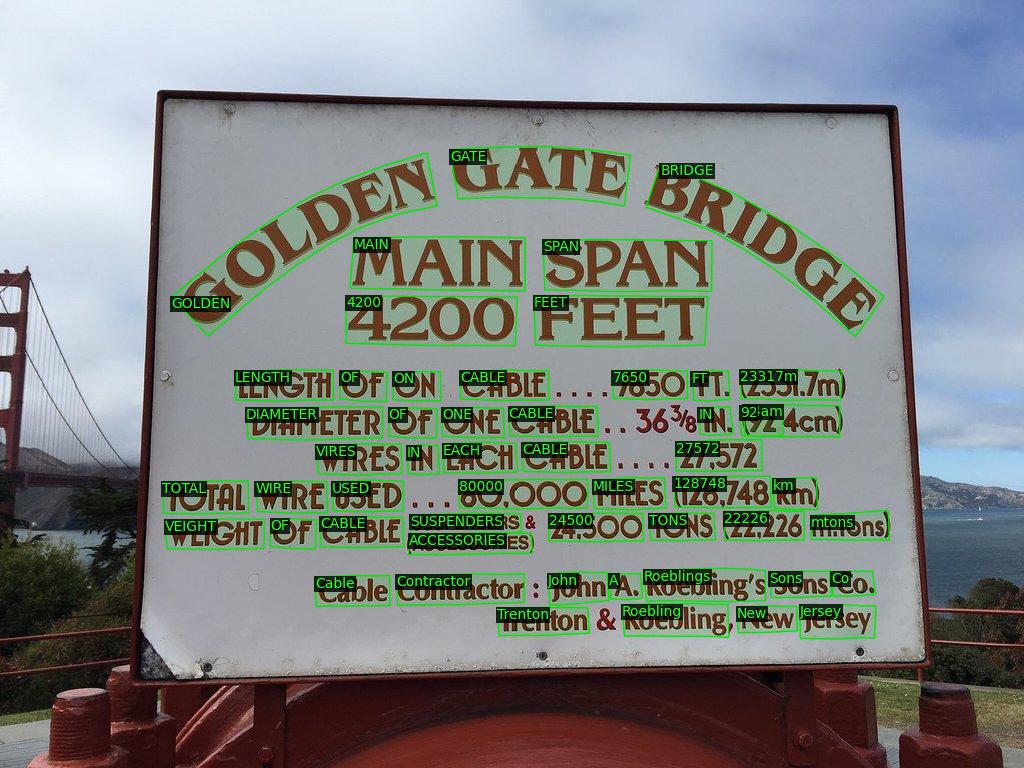}\\
(e)&(f)&(g)&(h)\\
\includegraphics[height=2.5cm, width= 2.5cm]{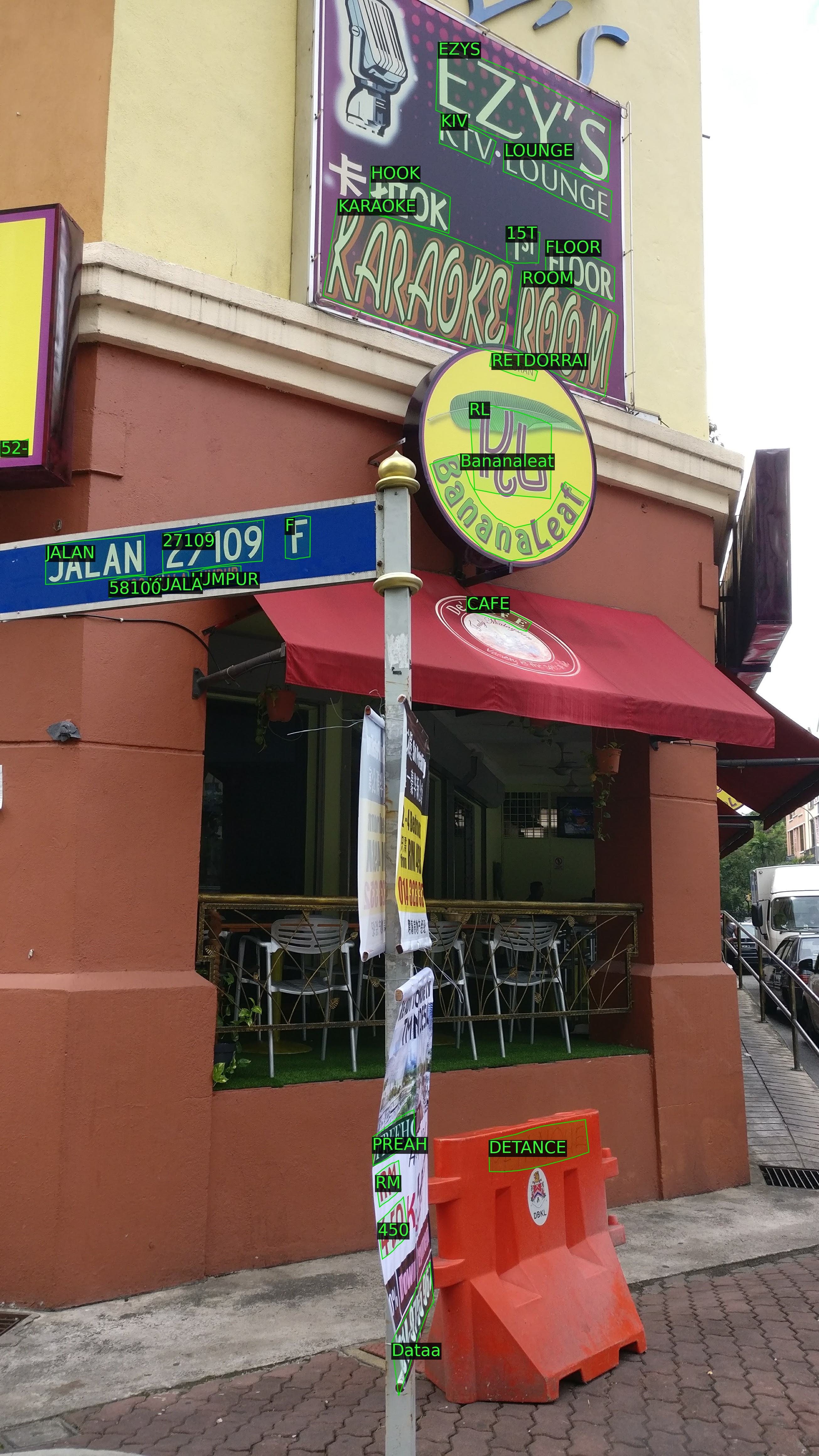}&
\includegraphics[height=2.5cm, width= 2.5cm]{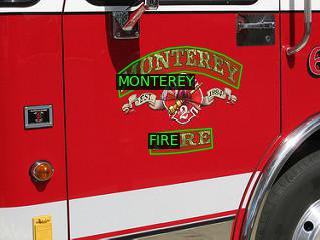}&
\includegraphics[height=2.5cm, width= 2.5cm]{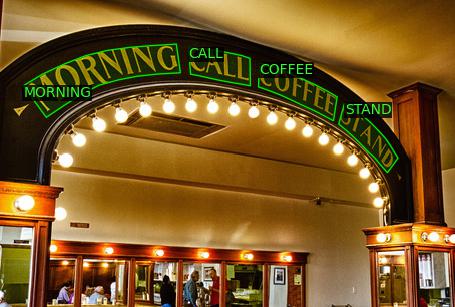}&
\includegraphics[height=2.5cm, width= 2.5cm]{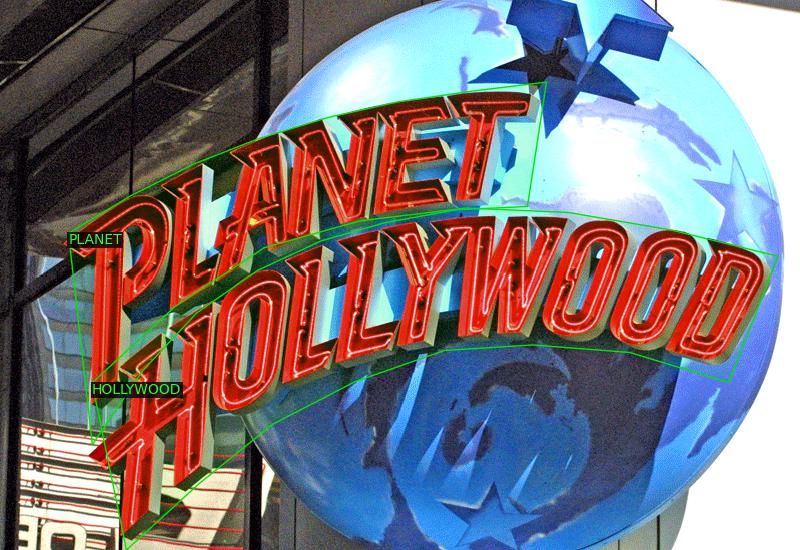}\\
(i)&(j)&(k)&(l)\\
\end{tabular}
\caption{\textbf{Performance Analysis on TotalText.} Some qualitative test cases of our method on images from TotalText.}
\label{fig:figsuptt}
\end{figure*}

\paragraph{Qualitative Performance Analysis on Total-Text} As shown in Fig.~\ref{fig:figsuptt}, TotalText is an arbitrary-shaped text spotting dataset at the word-level. The examples show the complexity of different scene conditions to spot text, where DG-TextSpotter performs comprehensively well. Examples (a),(f) and (i) exclusively show images with lots of noises that it's hard to detect and on top recognise the text. DA-TextSpotter is robust and adapts itself to spot text in such complex scenes.  


\paragraph{Qualitative Insights on Regular Text ICDAR 2015} As shown in Fig.~\ref{fig:figsupic15}, the DA-TextSpotter framework is robust to spot even regular text and achieve superior performance. ICDAR-15 is a challenging benchmark as the examples in Fig.~\ref{fig:figsupic15} especially (e),(f), (g), (l) which do not have focused scene text and are highly challenging to spot them. DA-TextSpotter achieves state-of-the-art results even in this benchmark and the aforementioned qualitative results justify them. 

\paragraph{Scope for Improvement} As shown in Fig.~\ref{fig:fig9}, we observe some of the failure cases taken from the UWT dataset proposed in this work when compared to superior performance in other standard natural scene evaluation benchmarks~\cite{ch2020total,karatzas2015icdar, liu2019curved}. Results depict there is still scope for improvement in these scenarios especially when the text is not focused and the scene complexity is much higher.






\begin{figure*}[ht]
\centering
\begin{tabular}{cccc} 
\includegraphics[height=2.5cm, width= 2.5cm]{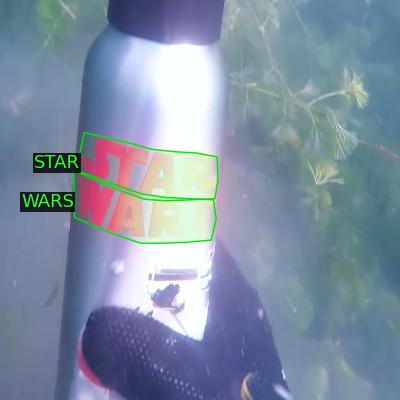}&
\includegraphics[height=2.5cm, width= 2.5cm]{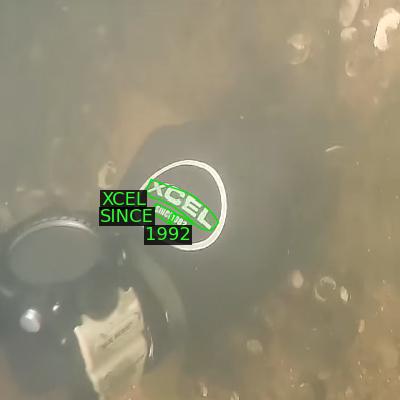}&
\includegraphics[height=2.5cm, width= 2.5cm]{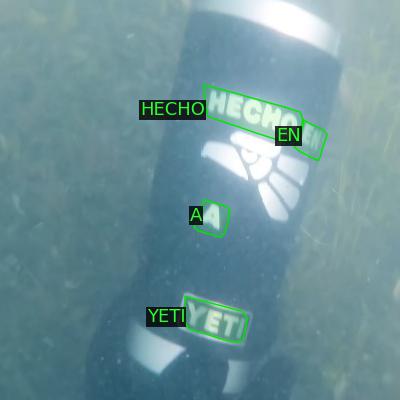}&
\includegraphics[height=2.5cm, width= 2.5cm]{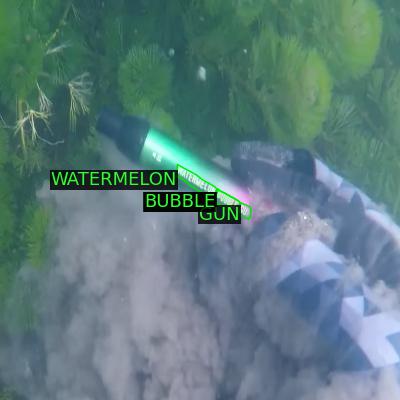}\\
(a)&(b)&(c)&(d)\\
\includegraphics[height=2.5cm, width= 2.5cm]{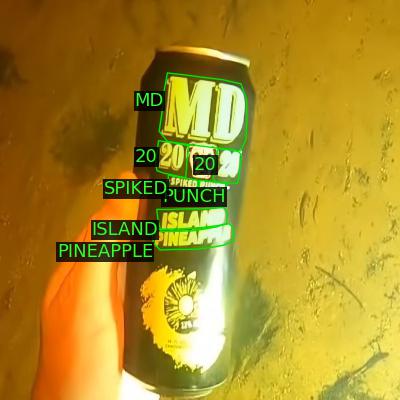}&
\includegraphics[height=2.5cm, width= 2.5cm]{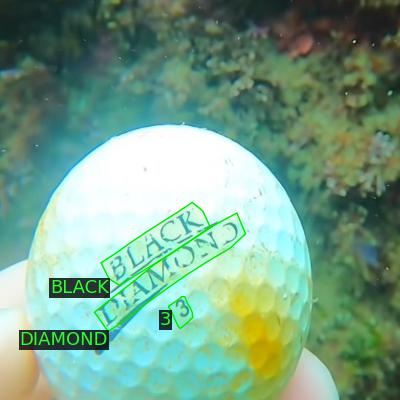}&
\includegraphics[height=2.5cm, width= 2.5cm]{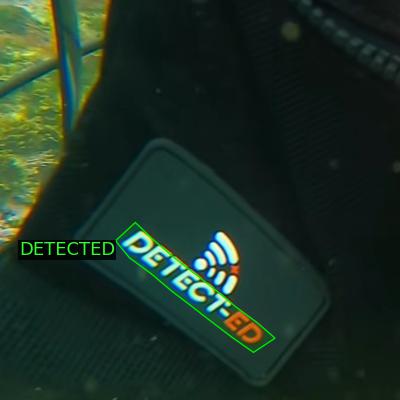}&
\includegraphics[height=2.5cm, width= 2.5cm]{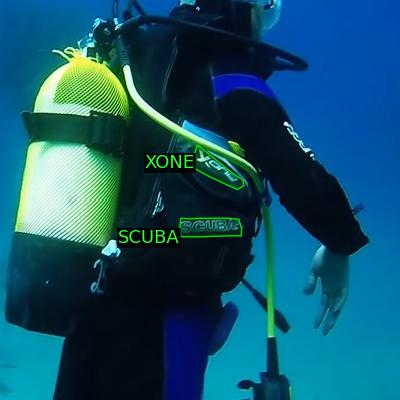}\\
(e)&(f)&(g)&(h)\\
\includegraphics[height=2.5cm, width= 2.5cm]{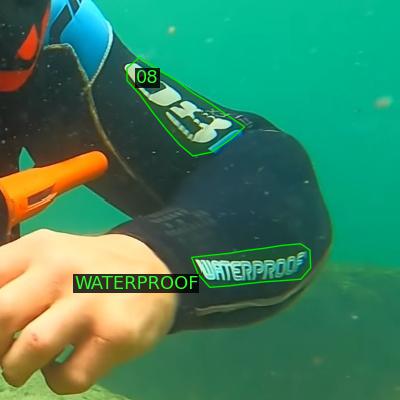}&
\includegraphics[height=2.5cm, width= 2.5cm]{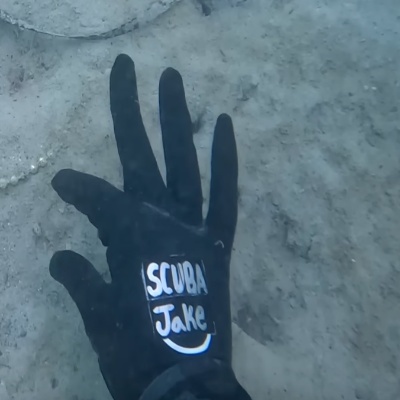}&
\includegraphics[height=2.5cm, width= 2.5cm]{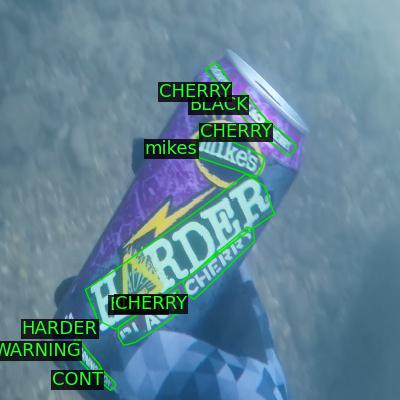}&
\includegraphics[height=2.5cm, width= 2.5cm]{images/248.jpg}\\
(i)&(j)&(k)&(l)\\
\end{tabular}
\caption{\textbf{Performance Analysis UWT.} Some qualitative test cases of our method on images from the UTSB dataset.}
\label{fig:figsupvin}
\end{figure*}

\begin{figure*}[ht]
\centering
\begin{tabular}{cccc} 
\includegraphics[height=2.5cm, width= 2.5cm]{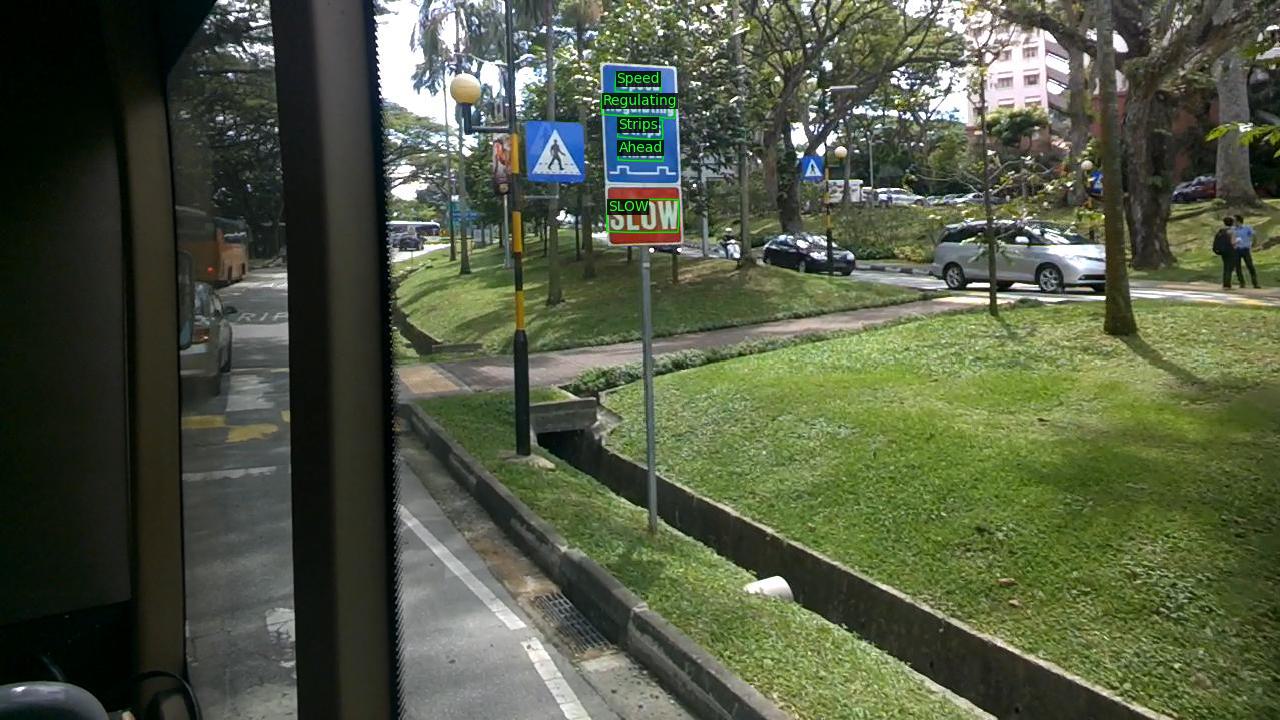}&
\includegraphics[height=2.5cm, width= 2.5cm]{images/img_10.jpg}&
\includegraphics[height=2.5cm, width= 2.5cm]{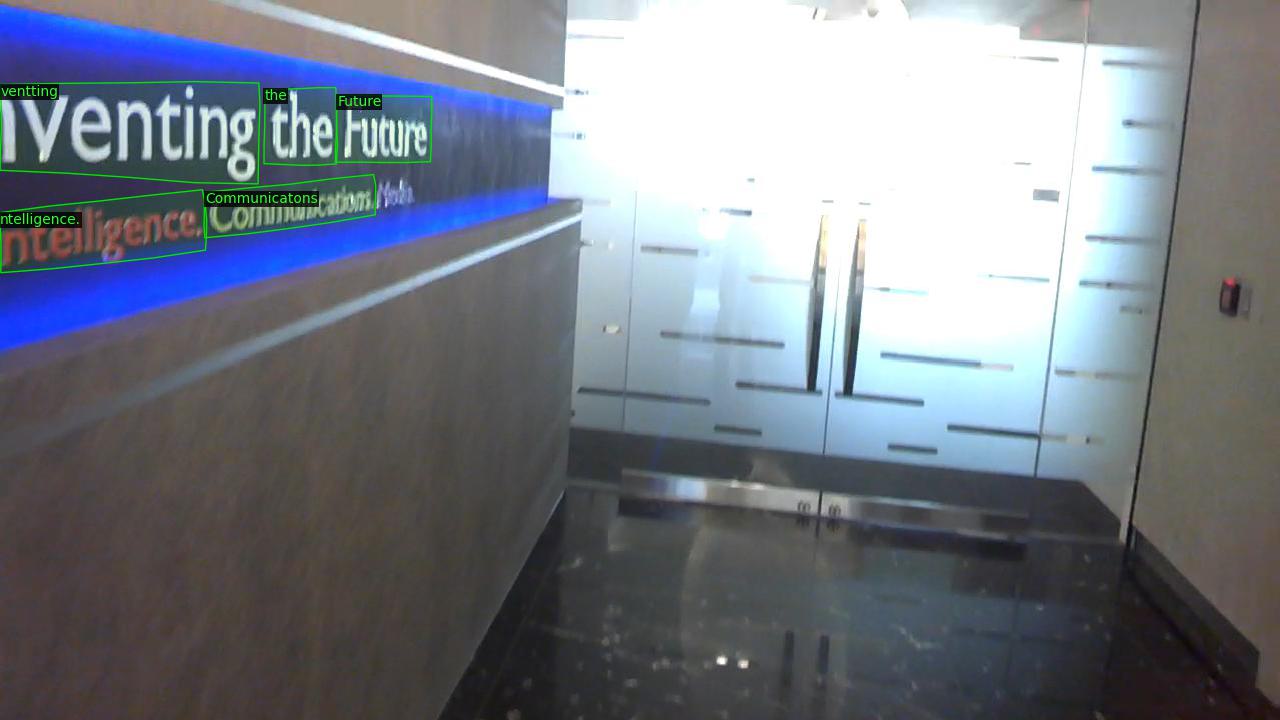}&
\includegraphics[height=2.5cm, width= 2.5cm]{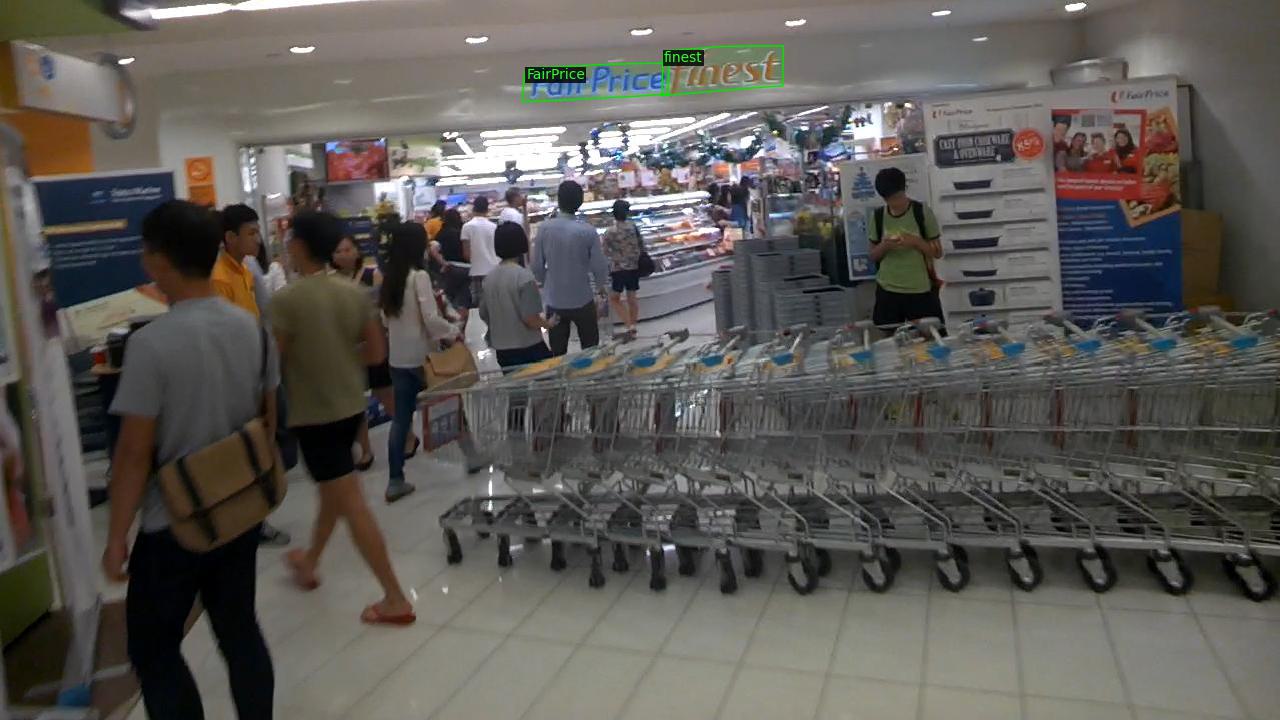}\\
(a)&(b)&(c)&(d)\\
\includegraphics[height=2.5cm, width= 2.5cm]{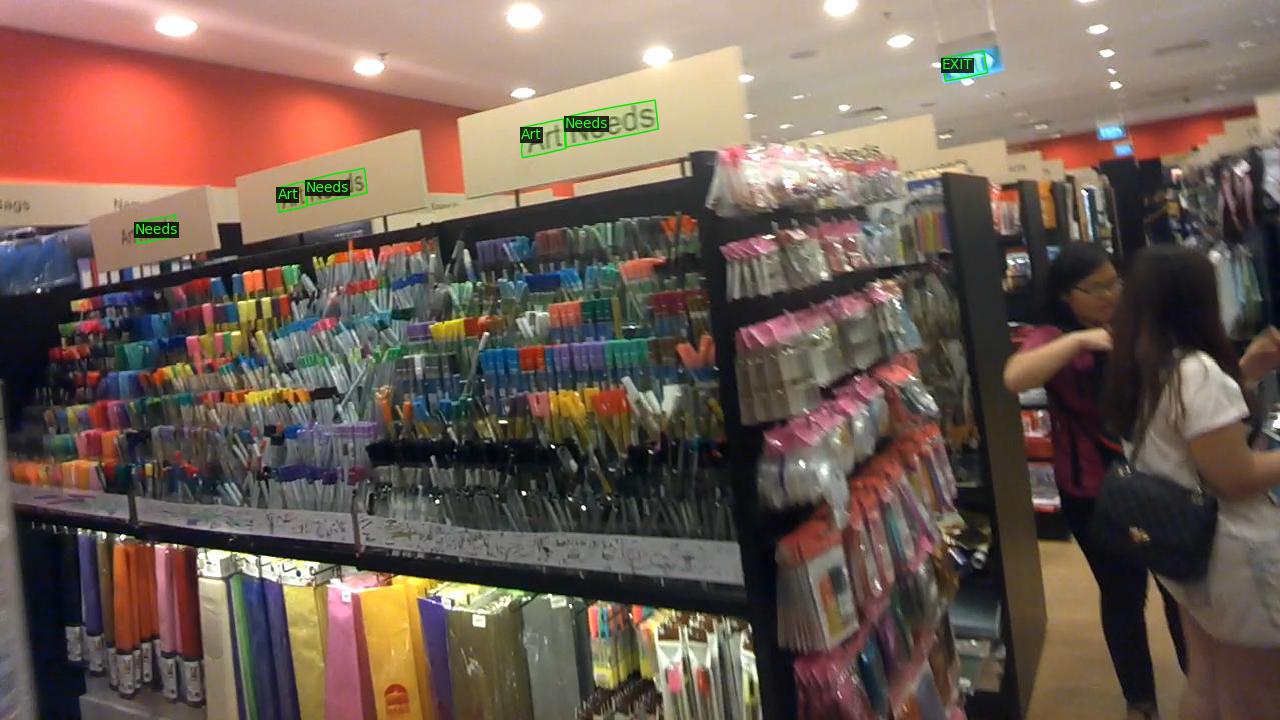}&
\includegraphics[height=2.5cm, width= 2.5cm]{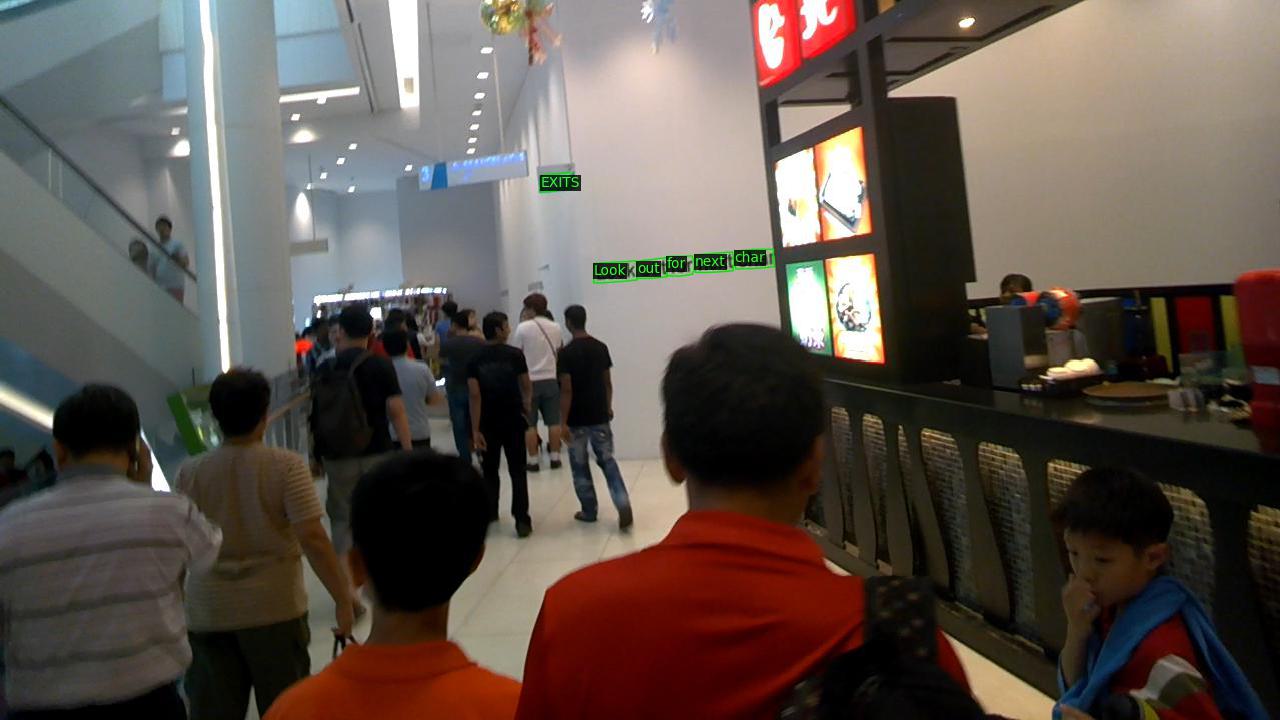}&
\includegraphics[height=2.5cm, width= 2.5cm]{images/img_57.jpg}&
\includegraphics[height=2.5cm, width= 2.5cm]{images/img_90.jpg}\\
(e)&(f)&(g)&(h)\\
\includegraphics[height=2.5cm, width= 2.5cm]{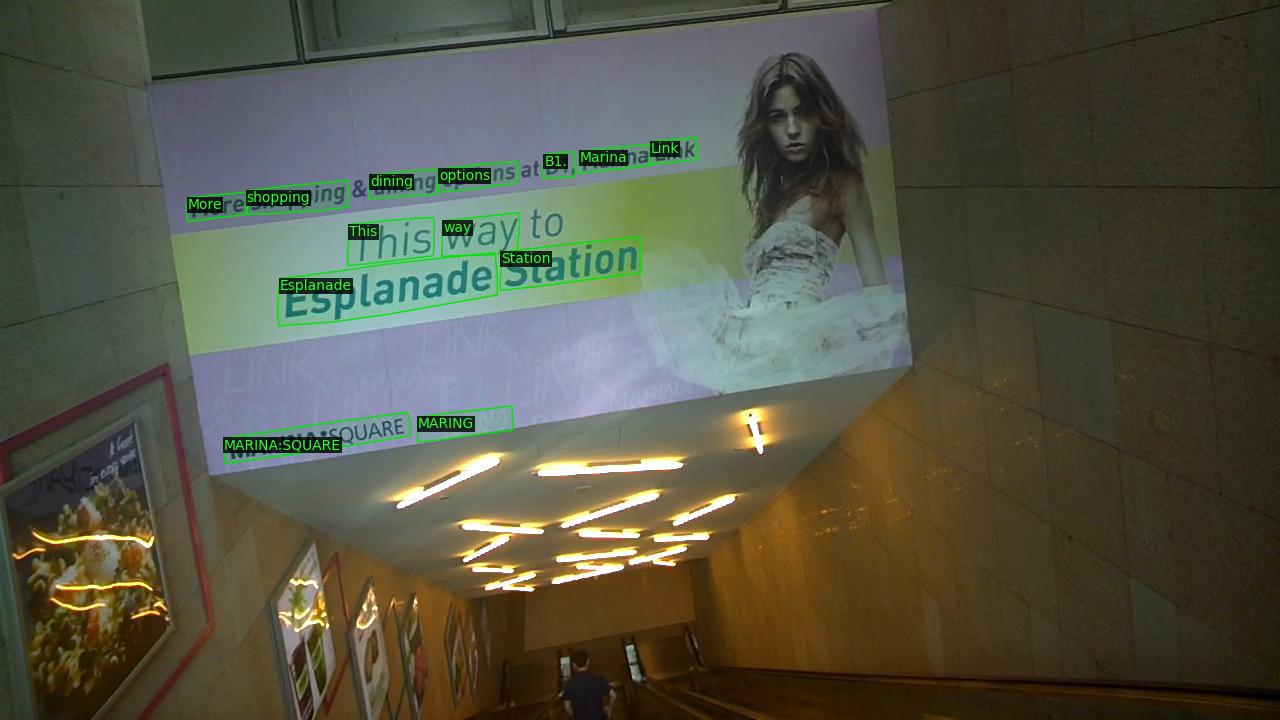}&
\includegraphics[height=2.5cm, width= 2.5cm]{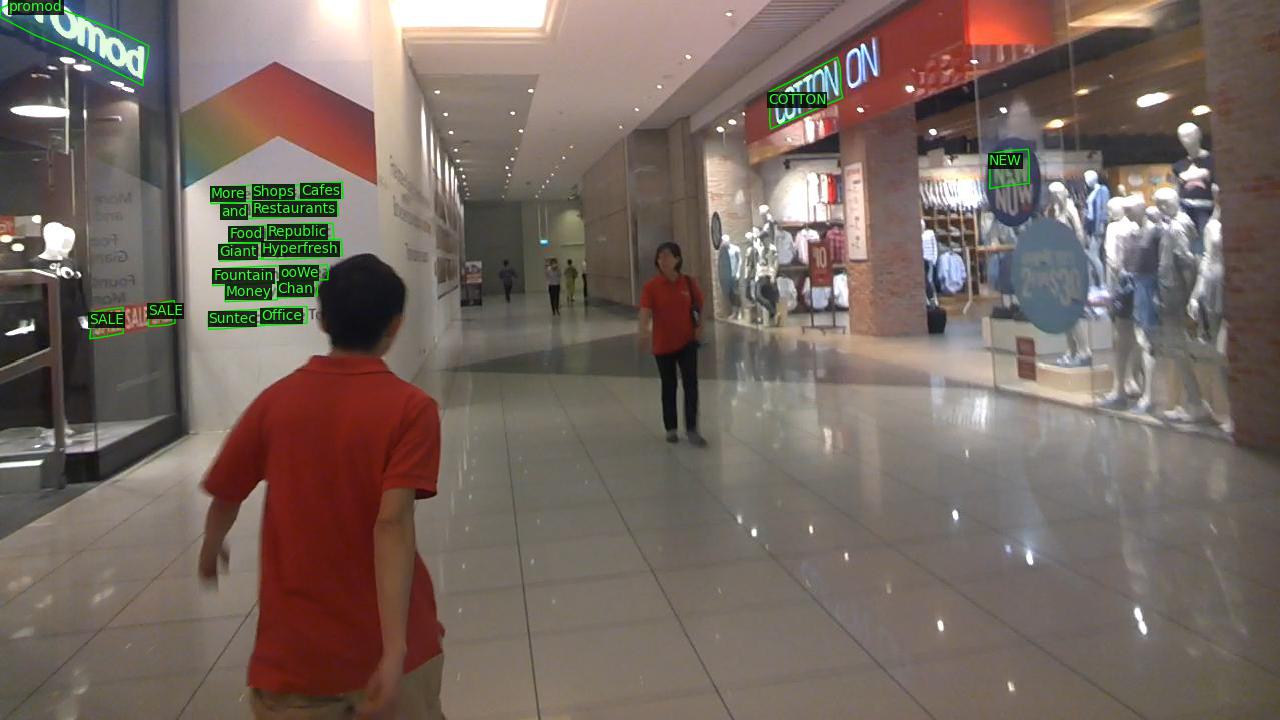}&
\includegraphics[height=2.5cm, width= 2.5cm]{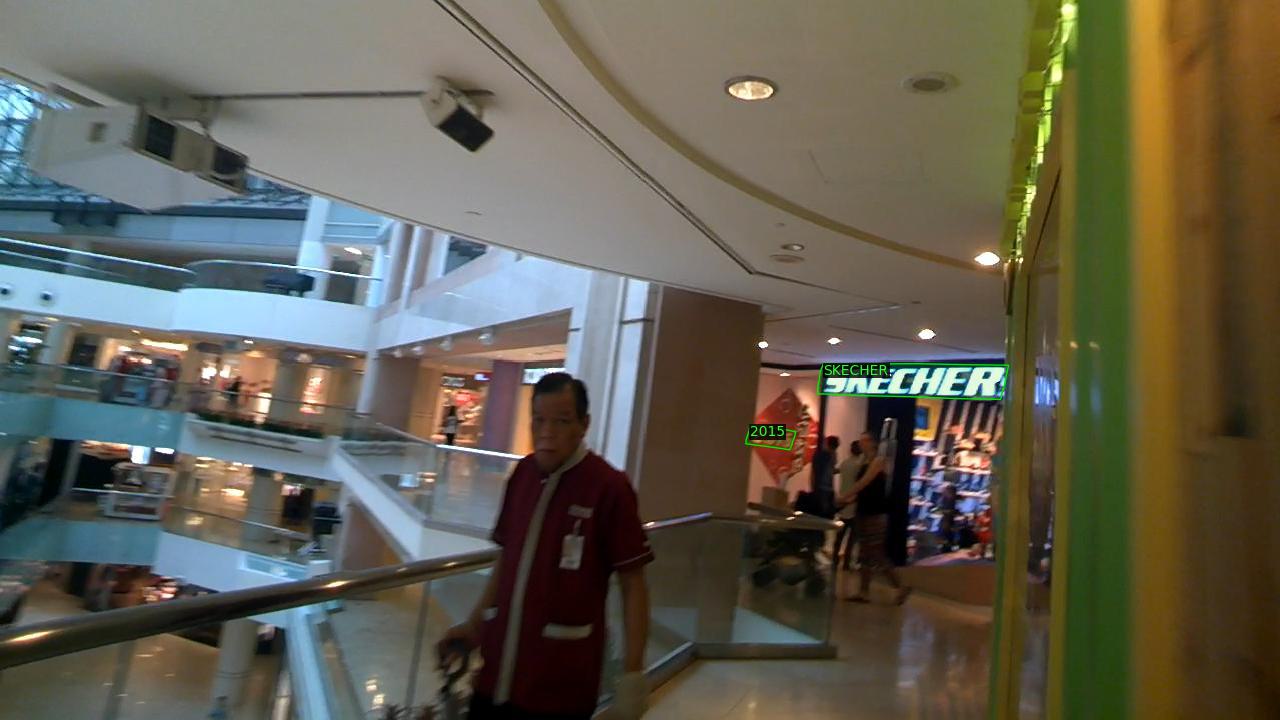}&
\includegraphics[height=2.5cm, width= 2.5cm]{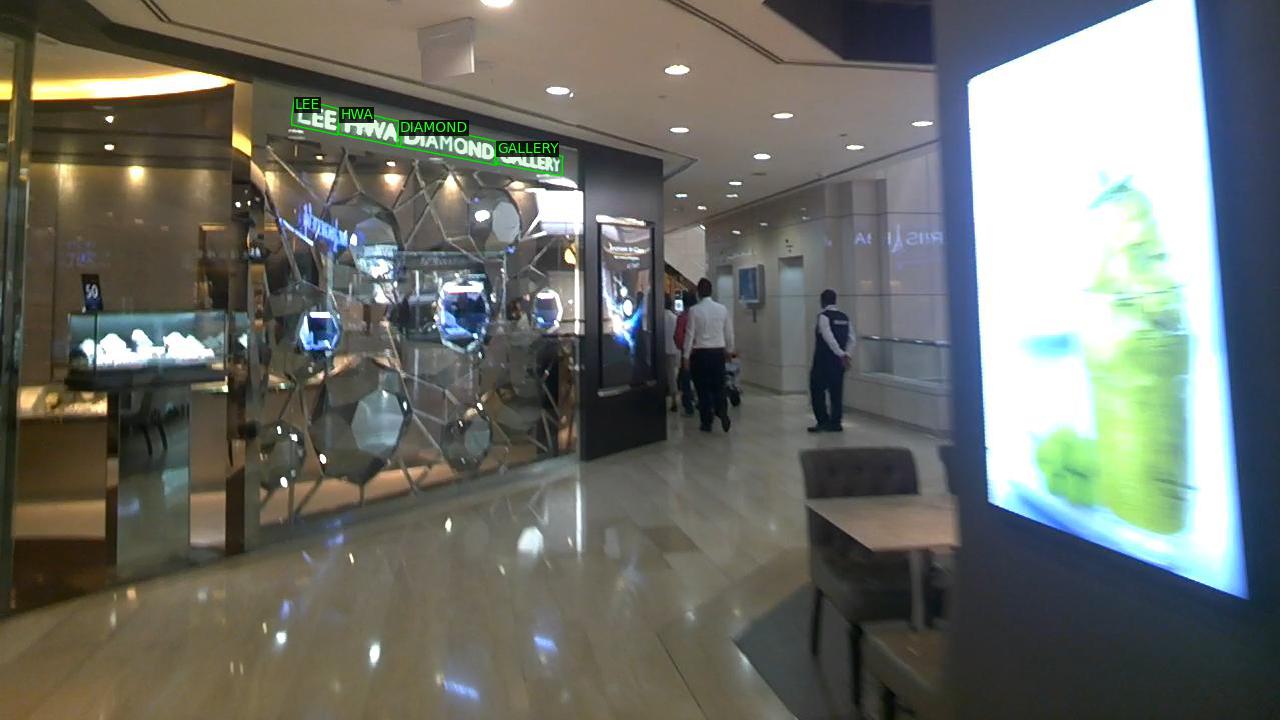}\\
(i)&(j)&(k)&(l)\\
\end{tabular}
\caption{\textbf{Performance Analysis on ICDAR15.} Some qualitative test cases of our method on images from ICDAR15.}
\label{fig:figsupic15}
\end{figure*}

{\small
\bibliographystyle{ieee_fullname}
\bibliography{egbib}
}